\documentclass[conference]{IEEEtran}
\usepackage{amsmath}  
\usepackage{amssymb}
\usepackage{times}
\usepackage[numbers]{natbib}
\usepackage{graphicx}
\usepackage{booktabs}
\usepackage{multicol}
\usepackage{multirow}
\usepackage{makecell}
\usepackage[nolist,nohyperlinks]{acronym}  
\usepackage[bookmarks=true]{hyperref}
\usepackage[capitalise]{cleveref}  
\usepackage{tikz}
\usepackage{pgfplots}
\pgfplotsset{compat=1.18}

\newcommand{\algname}{HumanFlow}

\usepackage{xifthen}
\usepackage{xparse}
\usepackage{subdepth}
\usepackage{leftidx}
\usepackage{bm}

\newcommand{\bbm}{\begin{bmatrix}}
\newcommand{\ebm}{\end{bmatrix}}

\DeclareMathAlphabet{\mbf}{OT1}{ptm}{b}{n}
\newcommand{\mbs}[1]{{\bm{#1}}} 

\newcommand{\mbsbar}[1]{{\overline{\boldsymbol{#1}}}}
\newcommand{\mbshat}[1]{{\hat{\boldsymbol{#1}}}}
\newcommand{\mbstilde}[1]{{\tilde{\boldsymbol{#1}}}}

\newcommand{\mbsdot}[1]{{\dot {\boldsymbol{#1}}}}

\newcommand{\mbfbar}[1]{{\overline{\mbf{#1}}}}
\newcommand{\mbfhat}[1]{{\hat{\mbf{#1}}}}
\newcommand{\mbftilde}[1]{{\tilde{\mbf{#1}}}}

\newcommand{\mbfdot}[1]{{\dot{\mbf{#1}}}}

\newcommand{\cframe}[1]{{\smash{\protect\underrightarrow{\mathcal{F}}_{#1}}}}

\DeclareMathAlphabet{\mathbfit}{OML}{cmm}{b}{it}
\newcommand{\homo}[1]{{\mathbfit{#1}}}

\newcommand{\mbfh}[1]{{\homo{#1}}}

\newcommand{\pos}[2]{\leftidx{_{#1}}{ \mbf r}{_{#2}}} 
\newcommand{\posh}[2]{\leftidx{_{#1}}{\mbfh r}{_{#2}}} 

\newcommand{\vel}[3]{\leftidx{_{#1}}{\mbf v}{\IfValueTF{#2}{_{#2#3\hspace{2pt}}}{}}} 
\newcommand{\veltilde}[3]{\leftidx{_{#1}}{\mbftilde v}{\IfValueTF{#2}{_{#2#3\hspace{2pt}}}{}}} 
\newcommand{\velbar}[3]{\leftidx{_{#1}}{\mbfbar v}{\IfValueTF{#2}{_{#2#3\hspace{2pt}}}{}}} 
\newcommand{\velhat}[3]{\leftidx{_{#1}}{\mbfhat v}{\IfValueTF{#2}{_{#2#3\hspace{2pt}}}{}}} 
\newcommand{\veldot}[3]{\leftidx{_{#1}}{\mbfdot v}{\IfValueTF{#2}{_{#2#3\hspace{2pt}}}{}}} 

\newcommand{\acc}[3]{\leftidx{_{#1}}{\mbf a}{\IfValueTF{#2}{_{#2#3\hspace{2pt}}}{}}} 
\newcommand{\acctilde}[3]{\leftidx{_{#1}}{\mbftilde a}{\IfValueTF{#2}{_{#2#3\hspace{2pt}}}{}}} 
\newcommand{\accbar}[3]{\leftidx{_{#1}}{\mbfbar a}{\IfValueTF{#2}{_{#2#3\hspace{2pt}}}{}}} 

\newcommand{\rotvel}[3]{\leftidx{_{#1}}{\mbs \omega}{\IfValueTF{#2}{_{#2#3\hspace{2pt}}}{}}} 
\newcommand{\rotveltilde}[3]{\leftidx{_{#1}}{\mbstilde \omega}{\IfValueTF{#2}{_{#2#3\hspace{2pt}}}{}}} 
\newcommand{\rotvelbar}[3]{\leftidx{_{#1}}{\mbsbar \omega}{\IfValueTF{#2}{_{#2#3\hspace{2pt}}}{}}} 
\newcommand{\rotvelhat}[3]{\leftidx{_{#1}}{\mbshat \omega}{\IfValueTF{#2}{_{#2#3\hspace{2pt}}}{}}} 
\newcommand{\rotveldot}[3]{\leftidx{_{#1}}{\mbsdot \omega}{\IfValueTF{#2}{_{#2#3\hspace{2pt}}}{}}} 

\newcommand{\T}[2]{\leftidx{}{\mbfh T}{_{#1#2\hspace{2pt}}}} 
\newcommand{\q}[3]{\leftidx{_{#3}}{\mbf q}{_{#1#2\hspace{2pt}}}} 

\acrodef{MAV}[MAV]{Micro Air Vehicle}
\acrodef{POMDP}[POMDP]{Partially Observable Markov Decision Process}
\acrodef{ORCA}[ORCA]{Optimal Reciprocal Collision Avoidance}
\acrodef{HJ}[HJ]{Hamilton--Jacobi}
\acrodef{RL}[RL]{Reinforcement Learning}
\acrodef{MPC}[MPC]{Model Predictive Control}
\acrodef{MLP}[MLP]{Multi-Layer Perceptron}
\acrodef{FiLM}[FiLM]{Feature-Wise Linear Modulation}

\begin{document}
\title{\algname{} -- Diffusion-Driven MAV Navigation Among Humans via Tightly-Coupled Motion Tracking, Forecasting, and Control}

\author{
\IEEEauthorblockN{Simon Schaefer$^{1,2,3,\dagger}$ \,\, Joshua Näf$^{4}$ \,\, Stefan Leutenegger$^{4}$}
\vspace{0.2em}
\IEEEauthorblockA{
$^{1}$Technical University of Munich\,
$^{2}$MCML\,
$^{3}$MIRMI\,
$^{4}$ETH Zurich}
}
\maketitle

\begin{abstract}
Robust and accurate perception of humans in their 3D scene context is essential for integrating robots into everyday environments. Existing approaches, however, often fail to predict plausible and accurate human motion estimates that are consistent with the surrounding scene, especially in the presence of heavy occlusions or partial visibility. This can limit both safety and efficiency for robotic operations.
We introduce \algname, a latent diffusion model that unifies human motion tracking and forecasting, conditioned on the 3D scene context. We show that our human motion model produces smooth and accurate predictions under challenging conditions, including heavy occlusions, and outperforms state-of-the-art methods in tracking accuracy while being significantly more efficient. Furthermore, we show how \algname{}'s latent space can be tightly coupled with control by conditioning a flow-matching-based, approximate MPC policy on these representations. We validate our policy in simulation with real human trajectories for \ac{MAV} social navigation, demonstrating superior navigation performance and remaining collision-free, even under partial observability of the human. 
\end{abstract}

\IEEEpeerreviewmaketitle
\section{Introduction}
Robots operating in human-populated environments must navigate safely and efficiently despite the inherent complexity and uncertainty of human behavior. Ensuring safety requires the ability to continuously track and forecast human motion, even under partial observability. While such a task is effortless for humans, it remains a fundamental challenge for robotic systems due to the highly dynamic, articulated, and context-dependent nature of human motion.

Consider a micro aerial vehicle navigating a crowded corridor: a person steps out from behind a pillar, only their shoulder and arm momentarily visible before the robot's own motion sweeps them toward the frame edge. Within milliseconds the planner must infer a full-body state, anticipate where this person is heading, and commit to an avoidance maneuver, all from a partial, rapidly changing observation.
Much of the existing literature on perceiving human motion struggles to address conditions that are common in robotic applications, including dynamic camera motion, severe occlusions, and limited human visibility from the robot’s viewpoint. Regression-based approaches offer real-time performance but often produce inaccurate or physically implausible motion estimates. In contrast, optimization-based methods are typically more accurate and robust to occlusions, yet are computationally too demanding for onboard deployment.
\setcounter{footnote}{1} 
\stepcounter{footnote}
\footnotetext{Corresponding author: simon.k.schaefer@tum.de}

\begin{figure}[t!]
\centering
\includegraphics[width=\linewidth, trim=50 50 50 50, clip]{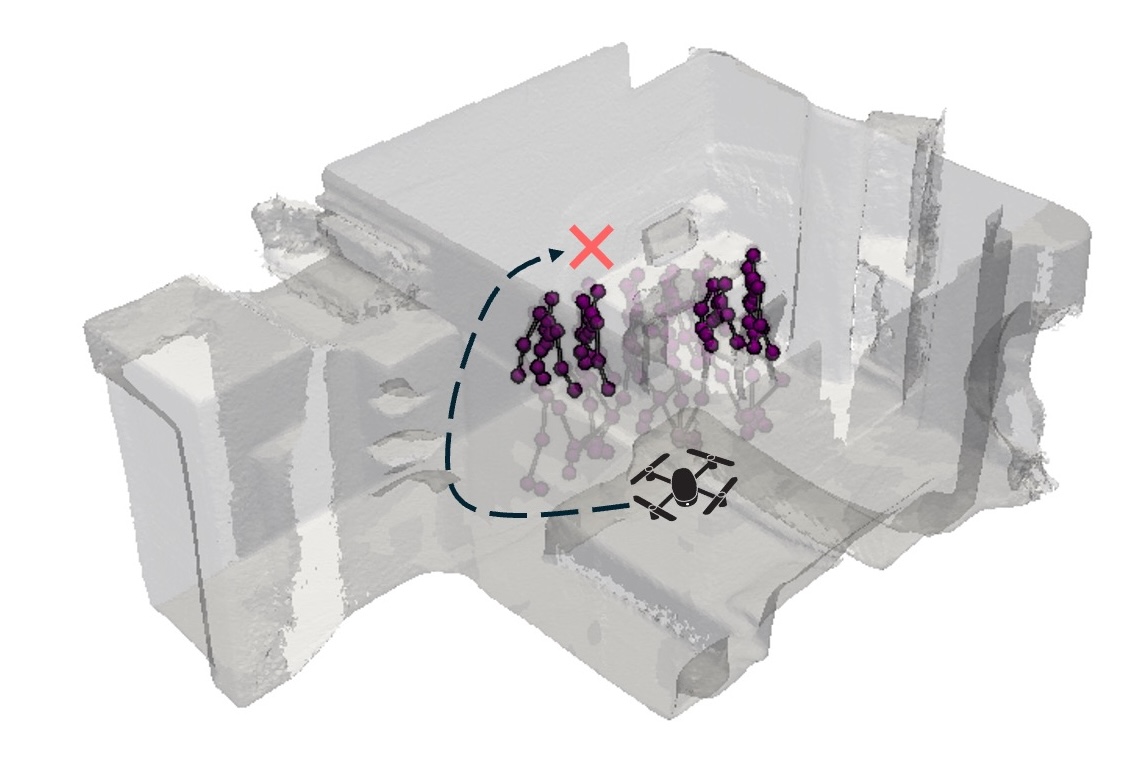}
\caption{An \ac{MAV} navigates a 3D environment containing a human whose motion is partially occluded by furniture. \algname{} efficiently tracks and forecasts human motion from noisy and heavily occluded observations while conditioning on the surrounding 3D scene using a latent diffusion model. Additionally, \algname{} compresses human and scene observations into a rich latent representation, enabling a computationally efficient flow-matching-based control policy. The resulting \ac{MAV} policy is collision-free and enables efficient navigation in cluttered environments despite severe occlusions.}
\label{fig:hero}
\vspace{-0.6cm}
\end{figure}

Neither approach incorporates the 3D scene context or has the ability to predict future motion, requiring the use of a separate forecasting model in a robotic use case.

In this work, we address these challenges by introducing \algname{}, a latent diffusion model for human perception in robotic settings. The model jointly tracks human motion from noisy and occluded observations and forecasts future motion, while being conditioned on the surrounding 3D scene context and operating in real time.
To demonstrate the effectiveness of the proposed approach, we apply \algname{} to the task of social navigation in which an \ac{MAV} must safely and efficiently navigate among humans. Social navigation in indoor environments exhibits many of the challenges discussed above, including dynamic human motion, rapidly changing viewpoints, and frequent occlusions (see \cref{fig:hero}).
We show that \algname{} is not only effective for human motion tracking and forecasting, but also learns a rich latent representation that facilitates learning robust \ac{MAV} policies. This enables safe social navigation under severe human occlusions while remaining computationally efficient for onboard deployment.

Our contributions are the following:
\begin{itemize}
    \item We propose \algname{}, a 3D context-conditioned latent diffusion model that jointly performs 3D human motion tracking and forecasting, capturing complex, multi-modal dynamics while remaining robust to noise and occlusions and efficiently representing long-horizon motion in a compact latent space.
    \item We develop a flow-matching-based \ac{MAV} policy that leverages the learned latent representation of \algname{} for proactive, human- and context-aware navigation, enabling real-time control while ensuring safety.
    \item We show the efficacy of \algname{} in several benchmarks. It produces smooth, realistic trajectories that align well with the 3D environment and achieves state-of-the-art accuracy. Furthermore, it is robust to high levels of noise and occlusions and is significantly more efficient than existing baselines.
    \item We demonstrate that using the latent representations of \algname{} enables our flow-matching-based \ac{MAV} policy to improve social navigation performance and to remain collision-free, even under challenging conditions with noisy or occluded observations.
\end{itemize}

\begin{figure*}
    \centering
    \includegraphics[width=\linewidth,trim=0.5cm 3.2cm 0.5cm 4cm,clip=true]{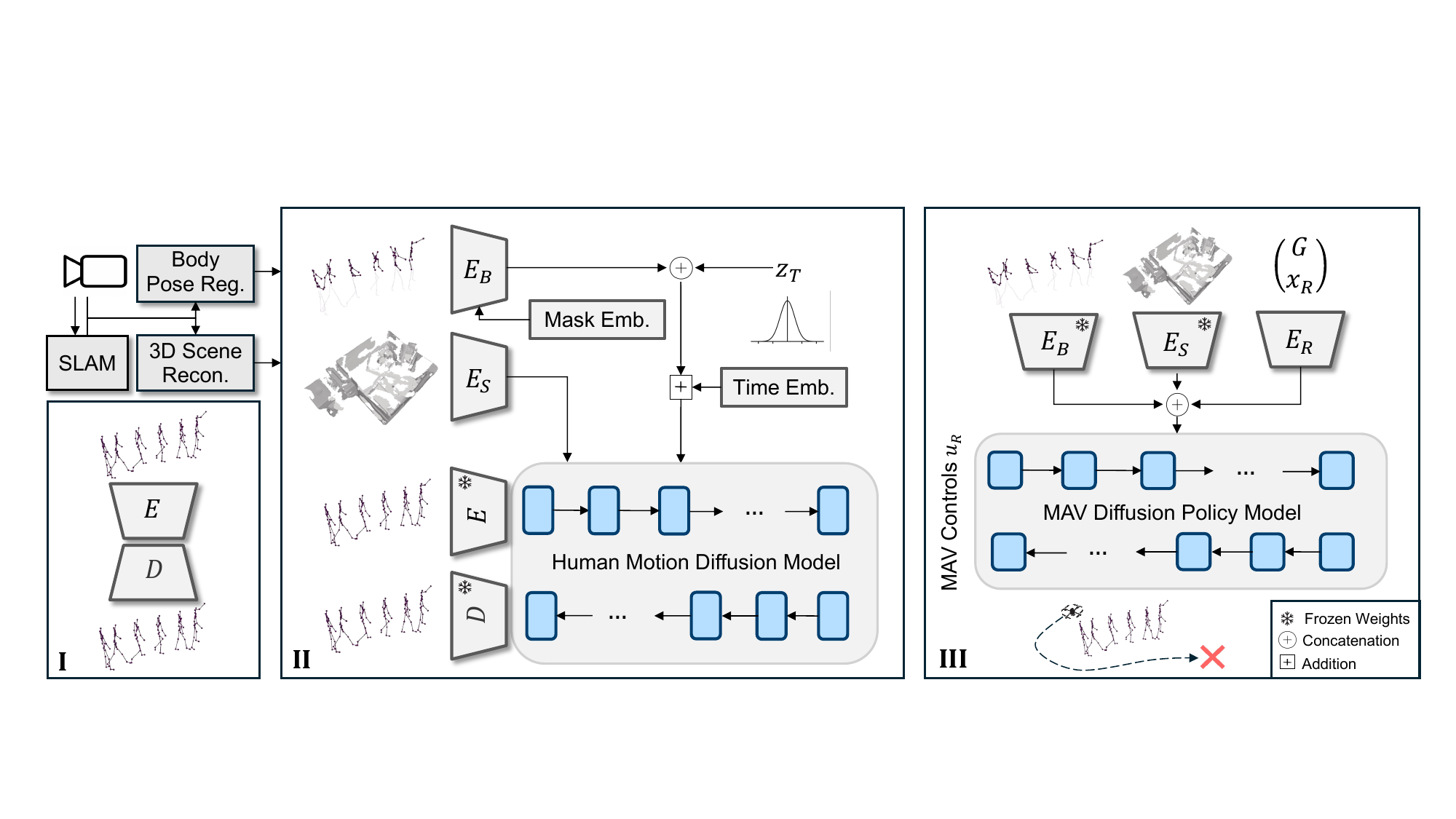}
    \caption{Overview of \algname{}. First, clean human motion is encoded into a compact latent space (I). Subsequently, a latent diffusion model is trained to denoise, complete, and forecast human motion sequences, conditioned on scene context and an initial estimate of visible human body frames obtained from video using SLAM and a per-frame human regression model (II). Finally, the latent human and scene embeddings, along with the \ac{MAV} state and goal, condition a flow-matching policy that outputs collision-free, goal-directed control commands (III).}
    \label{fig:architecture}
    \vspace{-0.5cm}
\end{figure*}

\section{Related Work} 
\subsection{Human Motion Tracking and Forecasting}
Prior work on 3D human pose and motion recovery largely falls into two camps: regression-based and optimization-based approaches. Regression methods \cite{khirodkar2022ochmr, kocabas2021pare, li2023jotr, zhang2023probabilistic, zhang2023nbf} predict pose or motion directly from images or video, and recent work has improved robustness to occlusion and noise \cite{zhang2023nbf, kocabas2021pare, li2023jotr}. Nevertheless, many regression models struggle with long-term occlusions and typically recover only local motion, which can produce jittery or implausible global trajectories. Optimization-based methods \cite{bogo2016simplify, rempe2021humor, shi2023phasemp, ye2023slahmr, zhang2021lemo} fit parametric body models \cite{loper2015smpl} using temporal priors ranging from handcrafted smoothness terms to learned motion models. These methods better match observations but are computationally expensive and sensitive to initialization, and rely on short-horizon or pairwise transition priors that break down under long occlusions.

Generative models have recently emerged as a powerful alternative. In particular, diffusion models \cite{chen2023executing, jiang2024motiongpt, karunratanakul2023gmd, tevet2023mdm, wang2023fgt2m} excel in synthesis \cite{tevet2023mdm, jiang2024motiongpt, karunratanakul2023gmd}, pose estimation \cite{foo2023hmdiff, gong2023diffpose, goel2023hmd2}, and forecasting \cite{curreli2025nonisotropic} because they can capture long-range spatio-temporal dependencies. RoHM~\cite{zhang2024rohm} applies coupled diffusion processes to iteratively model global trajectory and local motion, reconstructing globally consistent, complete motions from partial RGB(-D) observations under noise and occlusion. However, its iterative optimization over the full sequence is slow. By contrast, we compress motion into a latent space and employ a single diffusion model in a canonical human-centric space, yielding much faster inference while supporting simultaneous tracking and forecasting.

Scene-conditioned human motion models have mostly focused on forecasting \cite{cao2020longtermhuman, sun2026humof, xing2023sceneawarehuaman, hu2024hoimotion}, but typically assume a complete scene representation, which is unrealistic in real-time robotic settings. Xing et al.~\cite{xing2023sceneawarehuaman} enforce whole-body human–scene coherence by modeling mutual distances between the human and scene and predict future human–scene distance constraints to guide forecasting. This embeds motion into the scene but assumes noise-free input motion and full scene geometry and, as a discriminative model, produces a single deterministic trajectory despite the inherently multi-modal distribution of future motions. In contrast, our approach is generative, models the full distribution of human–scene interactions, explicitly distinguishes occupied, unoccupied, and unobserved scene regions to condition on incomplete scenes, and tolerates noisy past frames.


\subsection{Safe Navigation Among Humans}
Safe navigation among humans has been extensively studied for ground robots and, more recently, for aerial platforms. A central challenge is balancing accurate human motion reasoning with computational efficiency, particularly under uncertainty and partial observability.

Early work emphasized modeling human cooperation and intent to avoid overly conservative behavior. Trautman et al.~\cite{trautman2015robotcrowdnavigation} used interacting Gaussian processes to capture cooperative collision avoidance, mitigating the ``freezing robot problem'' in dense crowds. Sun et al.~\cite{sun2021trajectories} extended this idea by reasoning over distributions of human preferences rather than single trajectories, enabling richer representations of cooperative behavior. Bai et al.~\cite{bai2015intentionaware} formulated pedestrian interactions as a \ac{POMDP}, accounting for uncertainty in human intent, while Dragan et al.~\cite{dragan2017mathematicalmodels} argued for planning with both physical and cognitive models of humans.

Despite these advances, most cooperative or intention-aware methods assume simplified trajectory models or become computationally prohibitive in highly dynamic, multi-human environments. Safety-critical methods based on \ac{HJ} reachability~\cite{mitchell2005hjreachability, fridovichkeil2018planningfastslow} provide formal guarantees but scale poorly, often requiring simplified dynamics. Efficient multi-agent approaches, such as \ac{ORCA}~\cite{vandenberg2011orca, samavi2025sicnav}, rely on simplified human motion assumptions and are less effective when accurate individual predictions matter. Hybrid strategies that combine learned forecasting with reachability constraints~\cite{schaefer2020leveragingnngradients} improve realism but remain computationally expensive and are typically restricted to 2D navigation.

Learning-based approaches, including \ac{RL} and inverse \ac{RL}~\cite{cheng2019endtoend, everett2021collision, kretzschmar2016sociallycompliant}, can implicitly capture human motion patterns and produce socially compliant policies. However, their safety guarantees are probabilistic, they generalize poorly under out-of-distribution conditions, and they are sample-inefficient. Applications to aerial robots~\cite{tallamraju2020aircaprl, salazar2024hierarchical} have explored drone navigation for human tracking or obstacle avoidance, but these approaches typically assume stationary humans, perfect observations, or static environments, limiting robustness in realistic, cluttered, and dynamic settings.

Recently, HumanHalo~\cite{schaefer2024humanmpc} proposed an MPC framework for human-aware \ac{MAV} navigation. Like our approach, it constrains only the initial control input $\mathbf{u}_0$ to prevent the \ac{MAV} reachable set from fully overlapping with the human reachable set. This avoids overly conservative trajectories while still guaranteeing safety under the assumption of perfect knowledge of the human’s current state.

In contrast to relying on expensive optimization or simplified human dynamics models, our work focuses on tightly integrating predictive human motion latents into the control policy via a flow-matching-based formulation. Leveraging these latent representations enables the policy to reason about complex, scene-specific 3D human dynamics without requiring the policy itself to learn these dynamics explicitly, thereby reducing the required model capacity.
This enables real-time \ac{MAV} control, achieving safe, smooth, and efficient navigation without assuming perfect knowledge of human motion or the environment.

\section{Preliminaries}
\subsection{Coordinate Frames, Transformations, and Notation}
Reference coordinate frames are written as $\cframe{A}$; points expressed in frame $\cframe{A}$ are $\pos{A}{}$. The homogeneous transform from $\cframe{B}$ to $\cframe{A}$ is $\T{A}{B}$, parameterized by position $\pos{A}{A B}$ and orientation quaternion $\q{A}{B}{}$. We use two primary frames: the world frame $\cframe{W}$ and the \ac{MAV} body frame $\cframe{B}$.

\subsection{MAV Model}
The \ac{MAV} state comprises its world-frame position $\pos{W}{B}$, orientation $\q{W}{B}{}$, and linear velocity $\vel{W}{}{}$:
\begin{align}
\mathbf{x} = [\pos{W}{B}, \q{W}{B}{}, \vel{W}{}{}]. \label{eq:state_definition}
\end{align}
The navigation frame $\cframe{N}$, used for control, is obtained by rotating $\cframe{W}$ about its $z$–axis by yaw $\psi$ and translating its origin to the \ac{MAV} position, giving:
\begin{equation}
\posh{N}{} = \T{N}{B} \posh{B}{} = [x \ y \ z \ 1]^\top,
\end{equation}
where $\posh{N}{}$ are homogeneous coordinates in $\cframe{N}$.

Following \cite{tzoumanikas2019linearmpc, darivianakis2014hybrid}, with thrust $\tau$ and under the ZYX Euler-angle convention (yaw $\psi\in[-\pi,\pi]$, pitch $\theta\in[-\pi/2,\pi/2]$, roll $\phi\in[-\pi,\pi]$), the \ac{MAV} dynamics are approximated by a second-order system:
\begin{subequations}
\begin{alignat}{2}
\ddot{x} &= g \theta - c_x \dot{x}, \\
\ddot{y} &= - g \phi - c_y \dot{y}, \\
\ddot{z} &= \tau - c_z \dot{z}, \\
\ddot{\theta} &= - b_1 \theta + b_2 \theta^r, \\
\ddot{\phi} &= - b_3 \phi + b_4 \phi^r,
\end{alignat}
\label{eq:model}
\end{subequations}
with continuous-time control inputs $\mathbf{u}_t = [\tau, \theta^r, \phi^r]$, reference orientations $\theta^r, \phi^r$, aerodynamic friction $c_x,c_y,c_z$, gravity $g$, and system-identification gains $c_i,b_i$. Under assumptions of symmetry, small attitudes, and negligible aerodynamic coupling (see \cite{tzoumanikas2019linearmpc, falanga2018pampc}), the system is effectively linear and yaw is controlled separately. We assume yaw can track the observed human body independently.
Discretizing \eqref{eq:model} with sampling time $T_s$ and horizon $T$ yields a time-invariant state-space form:
\begin{align}
\mathbf{x}_{k+1} &= \mathbf{A} \mathbf{x}_k + \mathbf{B} \mathbf{u}_k, \\
\mathbf{x}_{0:T} &= \mathbf{\Phi} \mathbf{x}_0 + \mathbf{\Gamma} \mathbf{u}_{0:T-1},
\label{eq:system_dynamics}
\end{align}
where $\mathbf{A},\mathbf{B}$ are the state-space matrices and $\mathbf{\Phi},\mathbf{\Gamma}$ their stacked forms.


\section{Context-Conditioned Human Motion Model}
Inspired by prior work on human motion modeling~\cite{tevet2023mdm,guzov2025hmd2}, we represent human motion using a diffusion model, which naturally captures multi-modal future behaviors, supports rich conditioning, and is robust to noisy and incomplete input data. 
Let $\mathbf{h}_{1:T}\in\mathbb{R}^{T\times J\times 3}$ denote a clean and complete sequence of $J$ 3D body joints, and let $\tilde{\mathbf{h}}_{1:T_{\text{obs}}}$ be a noisy and partially observed version with corresponding binary mask $\mathbf{m}_{1:T_{\text{obs}}}\in\{0,1\}^{T_{\text{obs}}\times J}$ indicating missing observations. 
Noisy human joints can be extracted efficiently using regression-based methods such as Multi-HMR~\cite{barabel2025multihmr} and visibility masks from standard 2D keypoint detectors. Scene context can be obtained via 3D reconstruction methods such as \cite{vespa2019supereight}. Note that our approach does not rely on completeness or accuracy of these inputs, making it robust to noisy and partial observations.

Given scene context $\mathcal{S}$, our objective is to model the conditional distribution:
\begin{equation}
p_\theta\!\left(\mathbf{h}_{1:T}\mid \tilde{\mathbf{h}}_{1:T_{\text{obs}}}, \mathbf{m}_{1:T_{\text{obs}}}, \mathcal{S}\right),
\end{equation}
which jointly denoises and completes the observed motion while forecasting plausible future trajectories for $T>T_{\text{obs}}$.

\textbf{Autoencoder.} 
To efficiently model long temporal horizons, we perform diffusion in a learned latent space. A temporal autoencoder consisting of an encoder $E(\cdot)$ and decoder $D(\cdot)$ maps clean motion sequences to a compact latent representation $\mathbf{z}\in\mathbb{R}^{d\times T'}$ with $T'\ll T$: 
\begin{equation}
\mathbf{z} = E(\mathbf{h}_{1:T}), 
\qquad 
\hat{\mathbf{h}}_{1:T} = D(\mathbf{z}).
\end{equation}

In addition to the standard L2 reconstruction loss on joint positions, we apply several regularization terms. A velocity loss encourages temporal smoothness by penalizing differences in predicted and ground-truth joint velocities. A bone length regularization ensures consistent skeletal structure by discouraging changes in bone lengths (see Eqs. (12)). The overall training loss is a weighted sum of all components:
\begin{equation}
\mathcal{L} = \mathcal{L}_{\text{MSE}} 
              + \lambda_{\text{vel}} \mathcal{L}_{\text{vel}}
              + \lambda_{\text{bone}} \mathcal{L}_{\text{bone}},
\end{equation}
where $\lambda_{(\cdot)}$ are weighting hyperparameters.

\textbf{Diffusion Model.} 
Extending prior work, we adopt a single unified model that performs both human motion tracking and forecasting. This is achieved through the use of mask embeddings, which indicate observed and unobserved portions of the input sequence and allow the same diffusion model to handle denoising, inpainting, and prediction. During training, random temporal masking is applied so that the same model learns to inpaint missing segments and to extrapolate future motion.

Noisy and incomplete inputs are encoded by a separate encoder $E_{B}(\cdot)$, which shares the same architecture as $E$. This design encourages the noisy-input encoder to map corrupted observations into a latent space aligned with the latent space of clean data, while allowing it to specialize in handling noise and missing joints during training. Missing joints are padded with the mean SMPL~\cite{loper2015smpl} joint configuration, and temporally masked segments are filled using zero-order hold along the sequence. The latent $\mathbf{z}_{\text{noisy}}$ is used as conditioning for the diffusion process by conditioning it on the noise, similar to \cite{ke2025marigold}.
\begin{equation}
\mathbf{z}_{\text{noisy}} = E_{B}(\tilde{\mathbf{h}}_{1:T_{\text{obs}}}).
\end{equation}

Scene-level context is incorporated through an additional encoder that operates on a human-centric occupancy grid representation of the environment. In contrast to prior approaches that assume a fully known scene~\cite{sun2026humof,cao2020longtermhuman,xing2023sceneawarehuaman}, we explicitly represent free, occupied, and unknown regions, following modern occupancy mapping frameworks~\cite{vespa2019supereight}. 
 A scene encoder $E_{\mathcal{S}}(\cdot)$ produces a global scene latent:
\begin{equation}
\mathbf{z}_{\mathcal{S}} = E_{\mathcal{S}}(\mathcal{S}).
\end{equation}
During training, we obtain incomplete scenes by randomly masking contiguous parts of the input scene. The encoded scene information is injected into the latent diffusion network via \ac{FiLM}~\cite{perez2018film}, providing global scene conditioning that biases the generated motion according to the surrounding geometry and uncertainty.

We train the diffusion model using clean state predictions (\textit{x0}-prediction) using DDPM~\cite{ho2020ddpm} with 100 denoising steps. During inference, we reduce the number of denoising steps to 10.
The overall training objective combines a standard diffusion denoising loss $\mathcal{L}_{\text{diff}}$ with several regularization terms similar to those used for the autoencoder. A velocity loss encourages temporal smoothness in the predicted motion. A foot-contact (skating) regularization penalizes unrealistic foot sliding by comparing predicted foot motion to contact labels. Finally, a bone length regularization ensures consistent skeletal structure. 
The total loss is a weighted sum of all components:
\begin{equation}
\mathcal{L} = \mathcal{L}_{\text{diff}} 
              + \lambda_{\text{vel}} \mathcal{L}_{\text{vel}}
              + \lambda_{\text{foot}} \mathcal{L}_{\text{foot}}
              + \lambda_{\text{bone}} \mathcal{L}_{\text{bone}},
\end{equation}
\begin{subequations}
\begin{align}
\mathcal{L}_{\text{vel}} &=  \frac{1}{J (T - 1)} \sum_k^{T-1} \sum_j^{J} ||\hat{\mathbf{v}}_{j, k} - \mathbf{v}_{j, k}||^2, \\
\mathcal{L}_{\text{foot}} &= \frac{1}{J_{\text{foot}} (T - 1)} \sum_k^{T-1} \sum_j^{J_{\text{foot}}} f_{jk} ||\hat{\mathbf{v}}_{j, k}||^2, \\
\mathcal{L}_{\text{bone}} &= \frac{1}{E (T - 1)} \sum_k^{T-1} \sum_j^J \sum_i^J \delta_{ij} ||e_{ij, k} - e_{ij, k-1}||^2,
\end{align}
\label{eq:loss-functions}
\end{subequations}
where $\lambda_{(\cdot)}$ are weighting hyperparameters, $\mathbf{v}_j$ the joint velocity of joint $j$ from numeric differentiation, $f_{jk}$ a foot contact label indicating whether joint $j$ is in contact with the ground at time $k$, $e_{ij}$ the bone length between joint $i$ and $j$, $\delta_{ij}$ indicating whether joint $i$ is the parent of joint $j$ in the skeleton, and $E$ the number of bones.
The motion encoders $E_{B}$ and $E$, the decoder $D$, and the denoising model are implemented as transformer encoders, while the scene encoder $E_{\mathcal{S}}$ consists of 3D convolutions followed by average pooling. For more details about the training data and model architectures, we refer to \cref{sec:experiments} and the supplementary materials.

\section{Human-Aware MAV Policy}
\label{seq:policy}
\algname{} learns a rich latent representation that can be directly leveraged for robot control. To demonstrate its effectiveness, we address the problem of human-aware navigation for an \ac{MAV}, where the objective is to compute a control policy that drives the \ac{MAV} toward a goal while avoiding collisions with both humans and the surrounding 3D environment. Formally, at each time step $k$, the policy $\Pi$ produces control inputs:
\begin{equation}
\mathbf{u}_{k} = \Pi(\mathbf{x}_0, \mathbf{g}, \mathcal{H}, \mathcal{S}),
\end{equation}
where $\mathbf{x}_0$ denotes the initial \ac{MAV} state, $\mathbf{g}$ the goal position, $\mathcal{H}$ the human motion, and $\mathcal{S}$ the scene. In contrast to prior approaches that rely on explicit parametric representations of human motion and scene geometry, we employ an implicit representation that captures complex human dynamics and human–scene interactions, by using the pretrained encoders $E_{B}$ and $E_{\mathcal{S}}$ to condition a flow-matching-based \ac{MAV} control policy.

To generate supervision for policy learning, we utilize the trained human motion diffusion model to sample plausible future human trajectories and solve an offline nonlinear trajectory optimization problem for the \ac{MAV}. This optimization yields collision-free, goal-directed trajectories that account for both the predicted distribution of human motion and the 3D scene structure.

Finally, the control policy is trained via flow matching to reproduce the distribution of optimized trajectories, conditioned on the \ac{MAV} state, goal, and latent human and scene representations. Operating in the learned latent space allows the policy to exploit rich information about human behavior and scene structure, enabling robust handling of complex human dynamics and severe occlusions while reducing computational complexity and simplifying policy learning. 

The full system architecture, including both the human motion model and the \ac{MAV} policy, is illustrated in \cref{fig:architecture}.

\textbf{Policy Formulation.}
Given an observed human motion sequence and scene context, we first extract latent representations using the pretrained human and scene encoders to produce $\mathbf{z}_{\text{noisy}}$ and $\mathbf{z}_{\mathcal{S}}$, respectively. Then, these latent features are flattened over time  and concatenated with the goal position expressed in the \ac{MAV} body frame, along with the \ac{MAV}’s initial velocity $\vel{B}{0}{}$, roll $\phi_0$, and pitch $\theta_0$. The resulting policy input is:
\begin{equation}
\mathbf{o} = \Big[ \mathrm{vec}(\mathbf{z}_{\text{noisy}}),\ \mathrm{vec}(\mathbf{z}_{\mathcal{S}}),\ \leftidx{{_B}}{\mathbf{g}},\ \vel{B}{0}{},\ \theta_0,\ \phi_0 \Big],
\end{equation}
where $\mathrm{vec}(\cdot)$ flattens the input over time.
By operating on these frozen latent features, the policy avoids explicit high-dimensional reasoning over human joint configurations and scene geometry. This enables implementation as a lightweight multilayer perceptron trained using flow matching, supporting efficient onboard inference while maintaining rich contextual awareness. More details on the model architecture can be found in the supplementary materials.

\textbf{Offline Trajectory Generation.}
Training data is generated offline using our human motion forecasting model. We sample real human motion sequences from the AMASS~\cite{naureen2019amass} and GIMO~\cite{zheng2022gimo} datasets. We use our human motion model to generate multiple plausible future human trajectories by masking out the last $T$ frames. For each rollout, we randomly sample an initial \ac{MAV} state $\mathbf{x}_0$ and a goal position $\mathbf{g}$ in the vicinity of the human.

\begin{figure}
    \centering
    \includegraphics[width=0.9\linewidth]{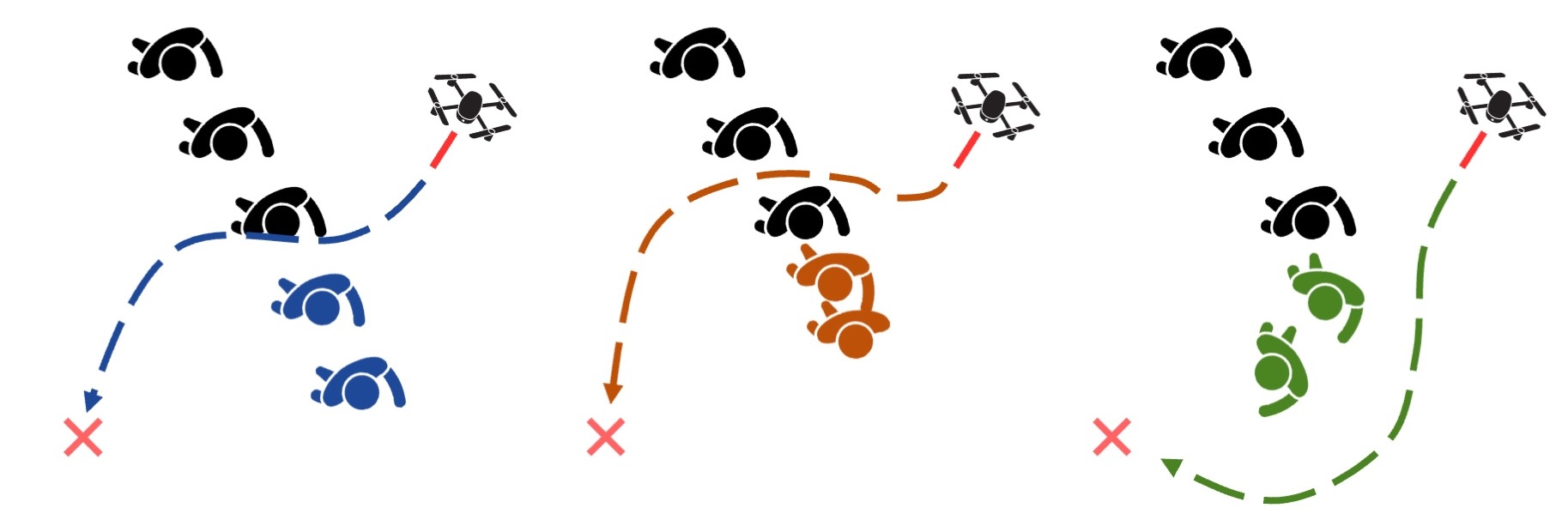}
    \caption{Exemplary offline ground-truth trajectory generation with three scenarios, each defined by a sampled human motion rollout. The initial control input (red segment) is shared across all scenarios, while the remaining control sequence is collision-free only within its respective scenario.}
    \label{fig:policy-gt-generation}
    \vspace{-0.7cm}
\end{figure}

We extend the idea presented in HumanHalo~\cite{schaefer2024humanmpc} to compute safe but not overly conservative \ac{MAV} controls by only guaranteeing safety for the initial control input $\mathbf{u}_0$ instead of the full horizon $\mathbf{u}_{0:T-1}$ (see \cref{fig:policy-gt-generation}).
While HumanHalo~\cite{schaefer2024humanmpc}, however, uses over-approximating reachable sets to represent the distribution of human motion $\mathcal{H}$, we use a set of 10 samples $\{\mathbf{h}^{0}_{1:T}, \mathbf{h}^{1}_{1:T}, \mathbf{h}^{2}_{1:T}, ... \}$ from the human motion model to capture the conditional distribution of future human motion, based on past motion and the scene.
Each rollout $i$ defines a scenario. We use a scenario MPC formulation to compute dynamically feasible and collision-free \ac{MAV} trajectories with respect to each scenario $\mathbf{h}^{i}_{1:T}$, where we enforce that the initial control action $\mathbf{u}_0$ is equal across all scenarios.
Since the optimized initial control action $\mathbf{u}^{*}_0$ is shared across all scenarios, and each optimized control sequence $\mathbf{u}^{*, i}$ is guaranteed to be collision-free in each scenario, $\mathbf{u}^{*}_0$ is guaranteed to be safe with respect to the full conditional human motion distribution.

For all scenarios we minimize the distance between the \ac{MAV}'s state $\mathbf{x}_k$ to the goal $\mathcal{G}$:
\begin{equation}
f(\mathbf{u}^i_{0:T-1}, \mathbf{x}^i_{0:T}) = \sum_{k=0}^{T} \left\| {\pos{W}{B}}_k - \mathbf{g} \right\|_2^2.
\end{equation}

We minimize the objective's expected value across all scenarios $i$. Formally, we solve the following optimization problem:
\begin{equation}
\begin{aligned}
\min_{\mathbf{u}^i_{0:T-1}} \quad & \mathbb{E}_{\mathcal{H}}[f(\mathbf{u}^i_{0:T-1}, \mathbf{x}^i_{0:T})] \\
\text{s.t.} \quad 
& \mathbf{x}_{k+1}^i = \mathbf{A} \mathbf{x}_k^i + \mathbf{B} \mathbf{u}_k^i, \quad \forall i \\
& \left\| {\pos{W}{B}}_k^i - \mathbf{h}^{i}_{j,k} \right\|_2 \ge d, \quad \forall j,\ \forall k,\ \forall i, \\
& \mathbf{u}_0^i = \mathbf{u}_0^j, \quad \forall i, j \\
& \text{ESDF}(\mathcal{S}, {\pos{W}{B}}_k^i) \geq d, \quad \forall i, \forall k \\
& \mathbf{u}_k \in \mathcal{U}, \quad \mathbf{x}_k \in \mathcal{X},
\end{aligned}
\end{equation}
where $\mathbf{h}^{i}_{j,k}$ denotes the position of the $j$-th human joint at time $k$ in the $i$-th scenario, $d$ is a predefined safety distance, and $\mathcal{U}$ and $\mathcal{X}$ denote control and state constraints. 
Whenever 3D context is given, we define $\text{ESDF}(\mathcal{S}, {\pos{W}{B}}_k^i)$ to be the euclidean signed distance function of the scene $\mathcal{S}$ at the position ${\pos{W}{B}}_k^i$, and we constrain it to be positive with safety margin $d$ across all scenarios.
The constrained non-linear optimization problem is solved using SLSQP~\cite{Kraft1988SQP}.

\textbf{Policy Learning via Flow Matching.}
Given the optimized \ac{MAV} trajectories, we train the policy using flow matching~\cite{lipman2023flowmatching}, inspired by \cite{julbe2025diffusionmpc}. Let $\mathbf{y}(t)$ denote a continuous-time interpolation between a base distribution $\mathbf{y}(0)$ and the target distribution $\mathbf{y}(1)$, parameterized by time $t\in[0,1]$. Flow matching learns a vector field $v_\theta(\mathbf{y}, t, \mathbf{o})$ such that:
\begin{equation}
\frac{d\mathbf{y}(t)}{dt} = v_\theta(\mathbf{y}(t), t, \mathbf{o}),
\end{equation}
and minimizes the objective:
\begin{equation}
\mathcal{L}_{\text{FM}} = \mathbb{E}_{t,\mathbf{y}(t)} \left[ \left\| v_\theta(\mathbf{y}(t), t, \mathbf{o}) - \frac{d\mathbf{y}(t)}{dt} \right\|_2^2 \right].
\end{equation}
In practice, $\mathbf{y}(1)$ corresponds to the optimized \ac{MAV} control sequence $\mathbf{u}_{0:T-1}^{*} \sim \mathbf{U}^{*}$. Since each offline scenario shares the same initial control $\mathbf{u}_0^{*}$, every optimized control sequence $\mathbf{u}_{0:T-1}^{*,i}$ from scenario $i$ is a sample from $\mathbf{U}^{*}$ and can be used to train the policy.

The policy is trained solely using the flow-matching loss $\mathcal{L}_{\text{FM}}$ on the generated ground-truth control sequences. At inference time, the learned policy produces control commands in real time. We found that 10 denoising steps are sufficient, enabling human-aware \ac{MAV} navigation that accounts for multi-modal human behavior while remaining computationally efficient for onboard deployment.

\begin{table*}[t]
\centering
\footnotesize
\begin{tabular}{c | c | l | c c c c c c | c}
\toprule
\multirow{2}{*}{Input} & 
\multirow{2}{*}{Noise} & 
\multirow{2}{*}{Method} & 
\multicolumn{3}{c}{GMPJPE $\downarrow$} & 
Accel$\downarrow$ & 
Skating$\downarrow$ & 
Dist$\downarrow$ & 
Runtimes$\downarrow$ \\
& & & -vis & -occ & -all & $[\text{m}/\text{s}^2]$ & [$\cdot$] & [mm] & [s]  \\
\midrule

\multirow{15}{*}{Occ-L} & \multirow{5}{*}{3} &
VPoser-t~\cite{pavlakos2019smplx} & 33.0 & 242.6 & 109.2 & 5.1 & 0.219 & 25.91 & 150 \\
&& HuMor~\cite{rempe2021humor} & 42.4 & 167.9 & 88.0  & 3.3 & 0.230 & 2.59 & 1800 \\
&& RoHM~\cite{zhang2024rohm} & \textbf{21.8} & \textbf{57.4} & \textbf{34.8} & \underline{2.3} & \textbf{0.078} & \textbf{0.69} & \underline{59} \\
&& Ours & \underline{27.6} & \underline{72.1}  & \underline{43.78}  & \textbf{0.58} & \underline{0.21} & \underline{1.55} & \textbf{0.22} \\[4pt]
\cline{2-10}\\[-4pt]

& \multirow{5}{*}{5} &
VPoser-t~\cite{pavlakos2019smplx} & 43.0 & 243.1 & 115.7 & 7.2 & 0.179 & 22.50 & 150 \\
&& HuMor~\cite{rempe2021humor} & 46.1 & 163.9 & 88.9  & 4.3 & 0.257 & 1.81  & 1800 \\
&& RoHM~\cite{zhang2024rohm} & \textbf{31.3} & \textbf{66.1}  & \textbf{44.0} & \underline{3.0} & \textbf{0.105} & \textbf{0.69} & \underline{59} \\
&& Ours & \underline{34.3} & \underline{80.2} &  \underline{50.99} & \textbf{0.60} & \underline{0.17} & \underline{1.42} & \textbf{0.22} \\[4pt]
\cline{2-10}\\[-4pt]

& \multirow{5}{*}{7} &
VPoser-t~\cite{pavlakos2019smplx} & 55.1 & 247.6 & 125.1 & 9.4 & 0.116 & 18.93 & 150 \\
&& HuMor~\cite{rempe2021humor} & 70.7 & 186.2 & 112.7 & 5.9 & 0.269 & 2.56  & 1800 \\
&& RoHM~\cite{zhang2024rohm} & \textbf{45.6} & \textbf{88.9}  & \textbf{61.3} & \underline{4.1} & \textbf{0.150} & \textbf{0.76} & \underline{59} \\
&& Ours & \underline{50.1} & \underline{92.2} & \underline{65.0}  & \textbf{0.64} & \underline{0.14} & \underline{1.00} & \textbf{0.22} \\

\midrule

\multirow{5}{*}{Occ-10\%} & \multirow{5}{*}{3} &
VPoser-t~\cite{pavlakos2019smplx} & 58.9 & 136.4 & 66.4  & 3.4 & 0.379 & 3.12 & 150 \\
&& HuMor~\cite{rempe2021humor} & 50.0 & 109.0 & 55.7  & 2.6 & 0.192 & 0.66 & 1800 \\
&& RoHM~\cite{zhang2024rohm} & \underline{26.3} & \underline{56.3}  & \underline{29.2} & \underline{2.3} & \textbf{0.085} & \underline{0.62} & \underline{59} \\
&& Ours & \textbf{17.5} & \textbf{40.6} & \textbf{19.8} & \textbf{0.48} & \underline{0.10} & \textbf{0.52} & \textbf{0.22} \\
\bottomrule
\end{tabular}
\caption{Scene-Free human motion tracking evaluation on AMASS~\cite{naureen2019amass}. Our model yields state-of-the-art tracking accuracy and physical plausibility while being significantly more computationally efficient than baselines.}
\label{tab:tracking-amass}
\vspace{-0.7cm}
\end{table*}

\section{Experiments}
\label{sec:experiments}
All experiments within this work have been conducted using the PyTorch framework \cite{pazke2019pytorch} on a single NVIDIA RTX A4000 GPU.
We used the AMASS~\cite{naureen2019amass} and GIMO~\cite{zheng2022gimo} datasets for both training and evaluation.

\textbf{AMASS}.
The AMASS dataset~\cite{naureen2019amass} provides large-scale, high-quality 3D human motion capture data with pose and shape annotations. We use the SMPL-X neutral body model and downsample all sequences to 10 fps. Following~\cite{zhang2024rohm}, TCD HandMocap, TotalCapture, and SFU are used for evaluation, with the remaining subsets reserved for training.

\textbf{GIMO}.
The GIMO dataset~\cite{zheng2022gimo} is a real-world human–scene interaction dataset. Human motion is captured using an IMU-based motion capture system, while 3D scenes are scanned with a LiDAR-equipped smartphone. GIMO uses the SMPL-X~\cite{pavlakos2019smplx} body model. It includes 11 subjects performing diverse actions, such as opening windows or lying in bed, across 19 scenes, resulting in 127  motion sequences comprising 192k frames.
We use a random train-test split, with 10\% of the data for testing and the remaining 90\% for training.

\textbf{Training Details}.
All models are trained with the Adam-W optimizer~\cite{loshchilov2018adamw}. We use a learning rate of 5e-4 and batch size 256 for the human motion autoencoder and diffusion model, and 4e-5 and batch size 64 for the policy model. All models are trained on sequences of 48 frames, which empirically balance sufficient context and computational cost.
Across model training we use $\lambda_{\text{vel}} = 1$, and $\lambda_{\text{foot}} = \lambda_{\text{bone}} = 0.1$.

We adopt a DDPM~\cite{ho2020ddpm} noise schedule with 100 training timesteps and 10 inference timesteps. The variance schedule is linear with $\beta$ values ranging from $10^{-4}$ to $2 \times 10^{-2}$. The model predicts denoised samples directly. Masked frames are replaced with the last valid frame before noise. When there is no last valid frame available, it is filled with the mean SMPL~\cite{loper2015smpl} joint configuration. This avoids introducing artificial discontinuities and serves as a good a priori estimate.

\textbf{Training Augmentations.} 
During training, we corrupt ground-truth SMPL-X~\cite{pavlakos2019smplx} motion sequences by applying noise  and masking. Gaussian noise is applied independently to body pose, global orientation, and translation parameters. Rotational parameters are converted to Euler angles, perturbed with zero-mean Gaussian noise with standard deviation $5^\circ$, and converted back to rotation vectors. Translation parameters are perturbed using zero-mean Gaussian noise with standard deviation $5 cm$. 
To simulate missing observations, we apply stochastic dropout augmentations. Entire frames are removed with probability $30\%$, and body-part-specific dropout is applied with probability $20\%$, masking contiguous body regions rather than individual joints. Additionally, with probability $80\%$, forecasting augmentation is applied by masking frames at the end of the sequence. The augmentation strength follows a cosine schedule over training iterations, gradually increasing the corruption difficulty.

For the MAV policy, each simulation episode involves random initial and goal states, so the outcome distribution itself is informative. The 25th, 50th, and 75th percentiles of time-to-goal are reported in \cref{tab:policy-evaluation-w-distribution}. Our method remains fully safe with tightly bounded trajectories across noise levels.

\begin{figure}
    \centering
    \setlength{\tabcolsep}{0pt}
    \begin{tabular}{c@{}c@{}c@{}}
        \footnotesize Conditioning &
        \footnotesize Prediction &
        \footnotesize Ground Truth \\
    
        \includegraphics[width=0.33\columnwidth, trim=250 100 200 150, clip]{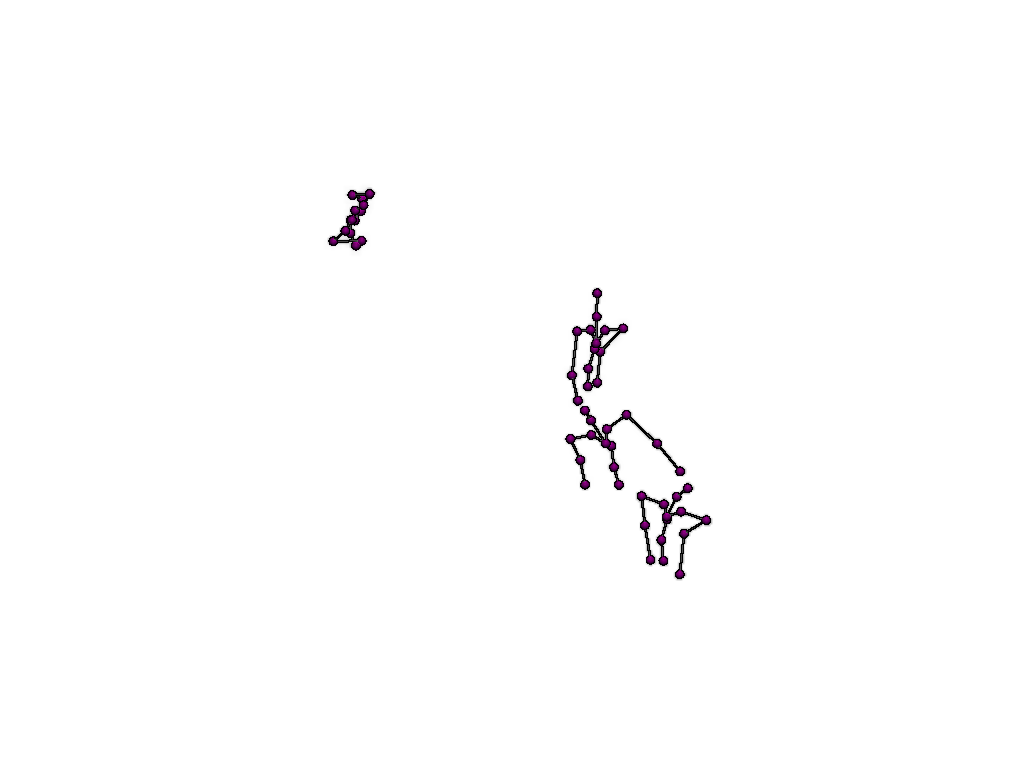} &
        \includegraphics[width=0.33\columnwidth, trim=250 100 200 150, clip]{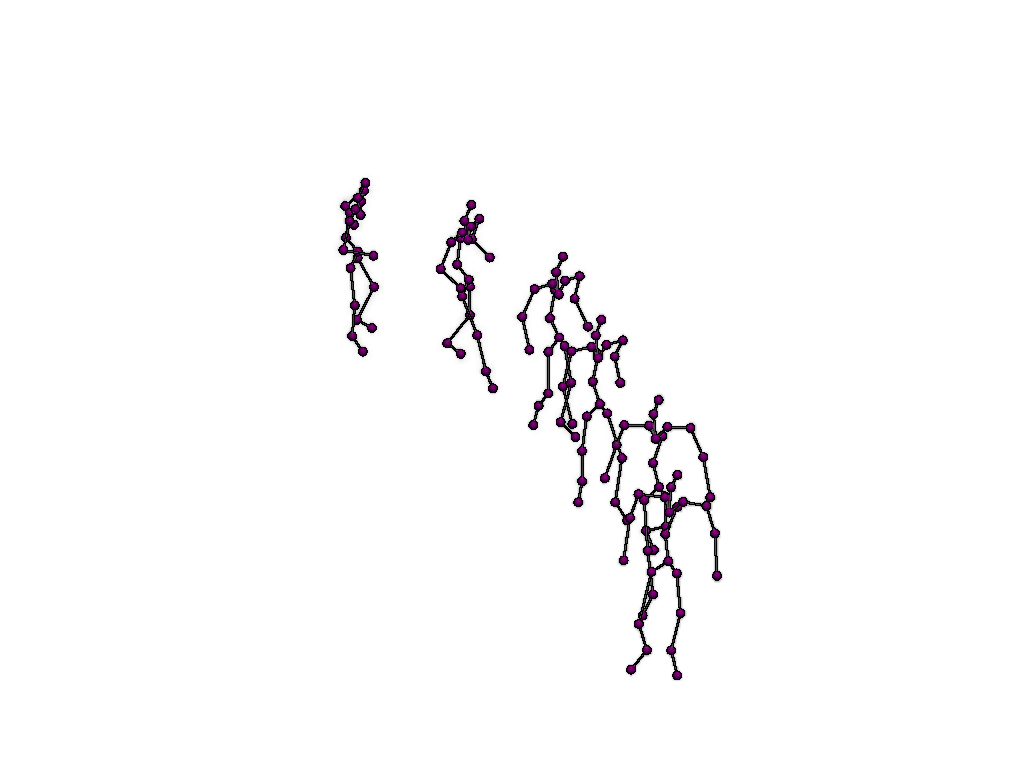} &
        \includegraphics[width=0.33\columnwidth, trim=250 100 200 150, clip]{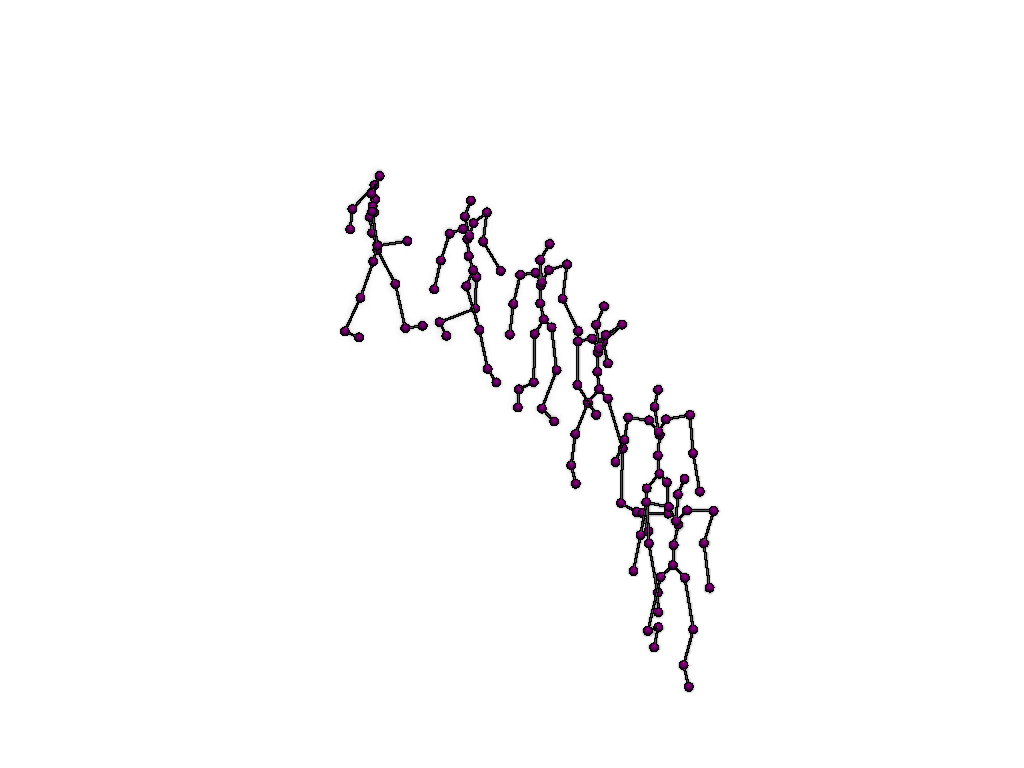} \\

        \includegraphics[width=0.33\columnwidth, trim=150 170 200 150, clip]{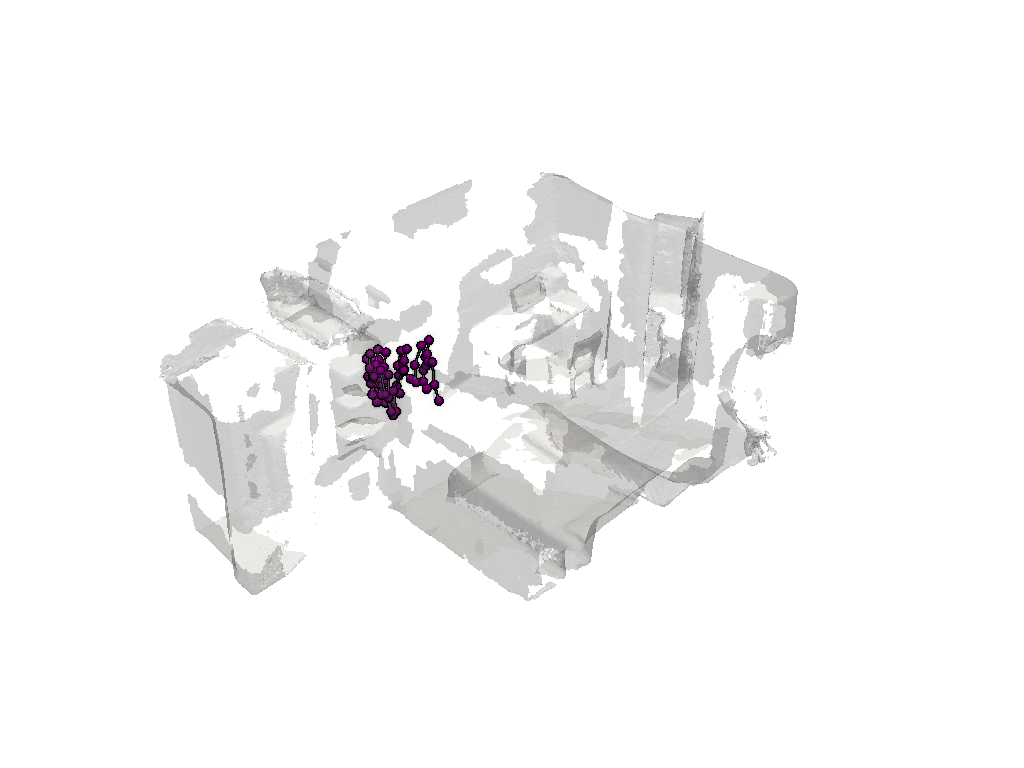} &
        \includegraphics[width=0.33\columnwidth, trim=150 170 200 150, clip]{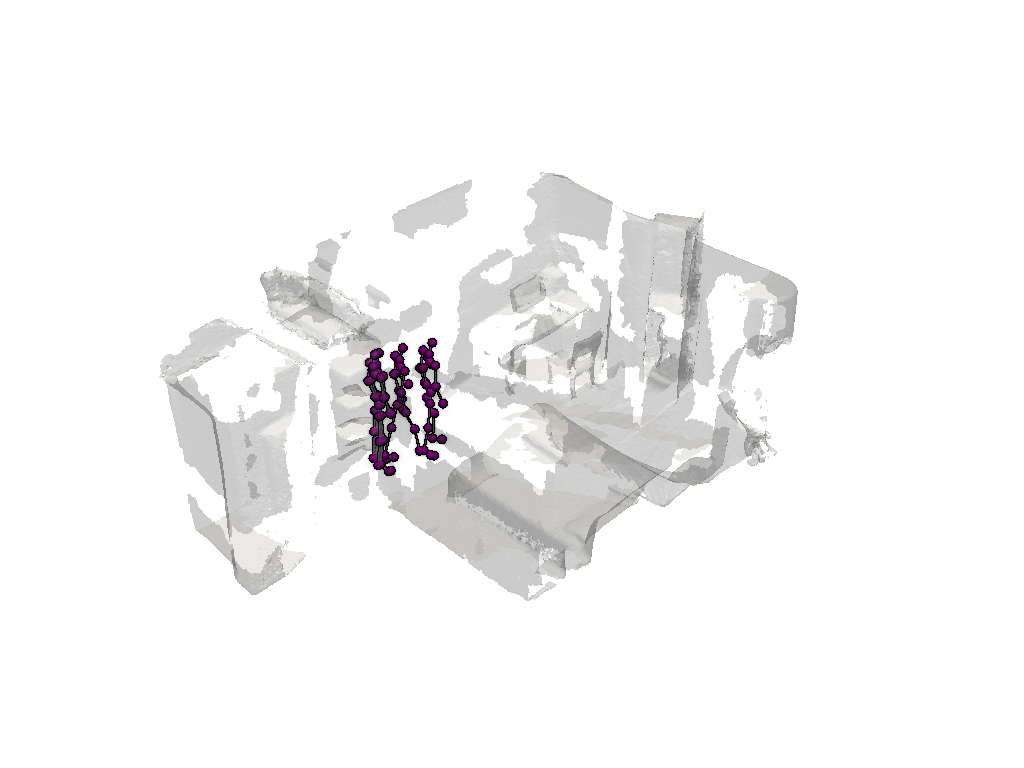} &
        \includegraphics[width=0.33\columnwidth, trim=150 170 200 150, clip]{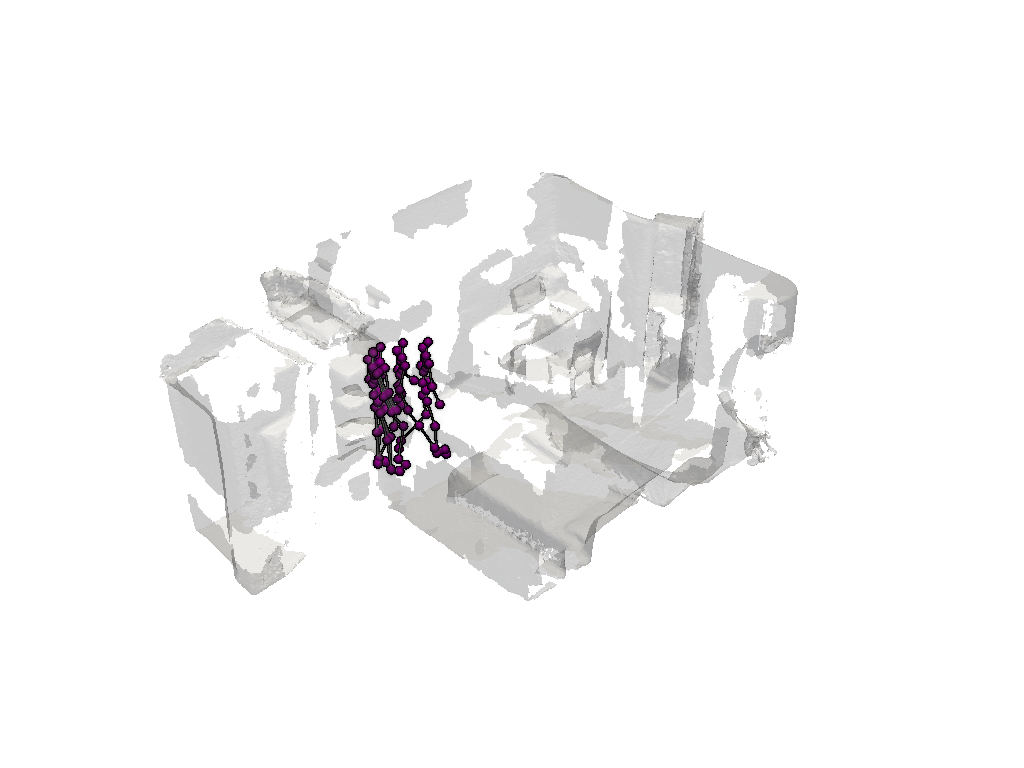} 

    \end{tabular}
    \caption{Qualitative examples of denoised and in-filled body motion with and without scene conditioning. (Upper) The human motion model is conditioned only on noisy observations of the upper body, while the lower body is occluded and several frames are entirely missing. (Lower) The model is conditioned on an incomplete scene. Only the upper body is observed and noisy. }
    \label{fig:tracking-qualitative}
    \vspace{-0.6cm}
\end{figure}

\subsection{Human Motion Tracking and Forecasting}
\label{seq:human-motion-eval}
\textbf{Metrics}.
We evaluate predicted body joint accuracy using mean-per-joint-position-error without alignment in global frame (GMPJPE). Metrics are reported separately for all, visible, and occluded joints, considering the 22 SMPL~\cite{loper2015smpl} body joints. In addition, we assess the physical plausibility of the body motion. We compute the acceleration error (Accel) as the difference in 3D joint accelerations between predictions and ground truth, measured in m/s². Foot skating (Skating) quantifies instances of foot sliding, defined following~\cite{zhang2021lemo} as cases where all foot joint velocities exceed 10cm/s, toe joints are less than 10 cm above the ground, and ankle joints are below 15 cm. Mean ground penetration (Dist) measures the extent of toe joint penetration into the ground, reported in millimeters, following \cite{rempe2021humor, shi2023phasemp, zhang2024rohm}. 

\textbf{Scene-Free Evaluation}.
To evaluate the robustness to noisy and occluded data, following \cite{zhang2024rohm}, we conduct two experiments: (1) motion denoising with lower-body infilling (Occ-L.) and (2) motion denoising with in-betweening (Occ-10\%). For a test sequence, (1) masks all lower-body joints to simulate common occlusions, while (2) masks a contiguous 10\% subsequence, requiring in-between motion generation. In both cases, Gaussian noise is added to SMPL-X~\cite{pavlakos2019smplx} parameters. Noise levels are defined as standard deviations of (k°, k°, k cm) for body pose, root orientation and translation, respectively, which accumulate along the kinematic tree.

We compare our method against several baselines specialized in recovering human motion from noisy and incomplete inputs (tracking scenarios). As shown in \cref{tab:tracking-amass}, our approach achieves state-of-the-art performance across all tests while being approximately 268× more efficient, as it requires no optimization in the loop and acts in a compressed latent space. This largely reduces the transformer's context length and therefore complexity. Notably, for Occ-10\%, our model is 47\% more accurate than RoHM~\cite{zhang2024rohm}, despite RoHM’s costly dual diffusion and optimization setup, highlighting our model’s ability to reason over in-between motion, which is a critical capability for safe \ac{MAV} navigation when humans are temporarily occluded and non-visible due to the \ac{MAV}'s ego-motion.

Moreover, by predicting body motion in a temporally compressed latent space, our model produces significantly smoother motion, substantially reducing joint acceleration errors across all test sets, by up to 6-fold. This also benefits the \ac{MAV} policy by preventing sudden jumps in predicted human motion, supporting safer and more stable navigation.

\begin{table}[!ht]
\centering
\footnotesize
\begin{tabular}{l | c c c c}
\toprule
Method & GMPJPE & Accel & Skat \\
\midrule
Ours wo/ Scene Encoder & 57.8  & \textbf{0.5} & 0.20 \\
Ours & \textbf{51.9}  & 0.6 & \textbf{0.12} \\
\bottomrule
\end{tabular}
\caption{Scene-Conditioned motion tracking evaluation on GIMO~\cite{zheng2022gimo} dataset. Conditioning on the scene boosts model accuracy as well as physical plausibility.}
\label{tab:tracking-gimo}
\vspace{-0.6cm}
\end{table}
\textbf{Scene-Conditioned Evaluation}.
Since the only prior method in this area, HMD2~\cite{guzov2025hmd2}, is not publicly available, we ablate our model’s performance with and without scene conditioning on the GIMO~\cite{zheng2022gimo} dataset. As shown in \cref{tab:tracking-gimo}, incorporating scene information improves tracking accuracy and reduces foot skating.

\textbf{Qualitative Examples}.
\cref{fig:tracking-qualitative} and \cref{fig:tracking-examples-supplementary} show qualitative results of our human motion denoising and infilling pipeline across diverse motions, including walking, jumping, sitting, and eating. Despite the variation in motion types, the generated sequences remain diverse while satisfying the imposed constraints, recovering smooth and temporally coherent full-body motion from noisy or occluded inputs, including missing-joint infilling and in-between motion generation.

\begin{table*}[ht!]
\centering
\setlength{\tabcolsep}{8pt} 
\renewcommand{\arraystretch}{1.2} 
\begin{tabular}{l | cc | cc}
\toprule
\textbf{Method}
& \makecell{\textbf{Collision}\\\textbf{Avoid. $\uparrow$ [\%]}} 
& \makecell{\textbf{Time To}\\\textbf{Goal $\downarrow$ [s]}} 
& \makecell{\textbf{Collision}\\\textbf{Avoid. $\uparrow$ [\%]}} 
& \makecell{\textbf{Time To}\\\textbf{Goal $\downarrow$ [s]}} \\
\midrule
& \multicolumn{2}{c|}{No Noise} & \multicolumn{2}{c}{Noise Level = 5} \\
\midrule
DC + Static Human Body & 86	& 4.12 & 84 & 5.25 \\
DC + Constant Velocity Model & 88 & 4.14 & 86 & 5.29\\
DC + Forecast-Based Model & 86 & 4.01 & 96 & 5.43 \\
\midrule
RSC + Forward Reachable Set & \textbf{100} & 4.50 & 90 & 5.61 \\
HumanHalo~\cite{schaefer2024humanmpc} & \textbf{100} & 4.46 & 92 & 5.47 \\
\midrule
Ours + Past Joints & 87 & 4.30 & 86 & 5.34 \\
Ours + Past \& Future Joints & 92 & 4.21 & 91 & 5.41 \\
Ours + Full Avoidance & \textbf{100} & 4.45 & \textbf{100} & 5.20 \\
\midrule
Ours & \textbf{100} & \textbf{3.80} & \textbf{100} & \textbf{3.91} \\
\bottomrule
\end{tabular}
\caption{Quantitative evaluation in simulation of the \ac{MAV} policy. Our policy remains safe across all sequences, even in the presence of high noise and occlusions, while yielding more efficient trajectories compared to the baselines.}
\label{tab:policy-evaluation}
\vspace{-0.7cm}
\end{table*}
\begin{figure}[!h]
    \centering
    \setlength{\tabcolsep}{0pt}
    \begin{tabular}{c@{}c@{}c@{}}
        \footnotesize Conditioning &
        \footnotesize Prediction (I) &
        \footnotesize Prediction (II) \\
    
        \includegraphics[width=0.3\columnwidth]{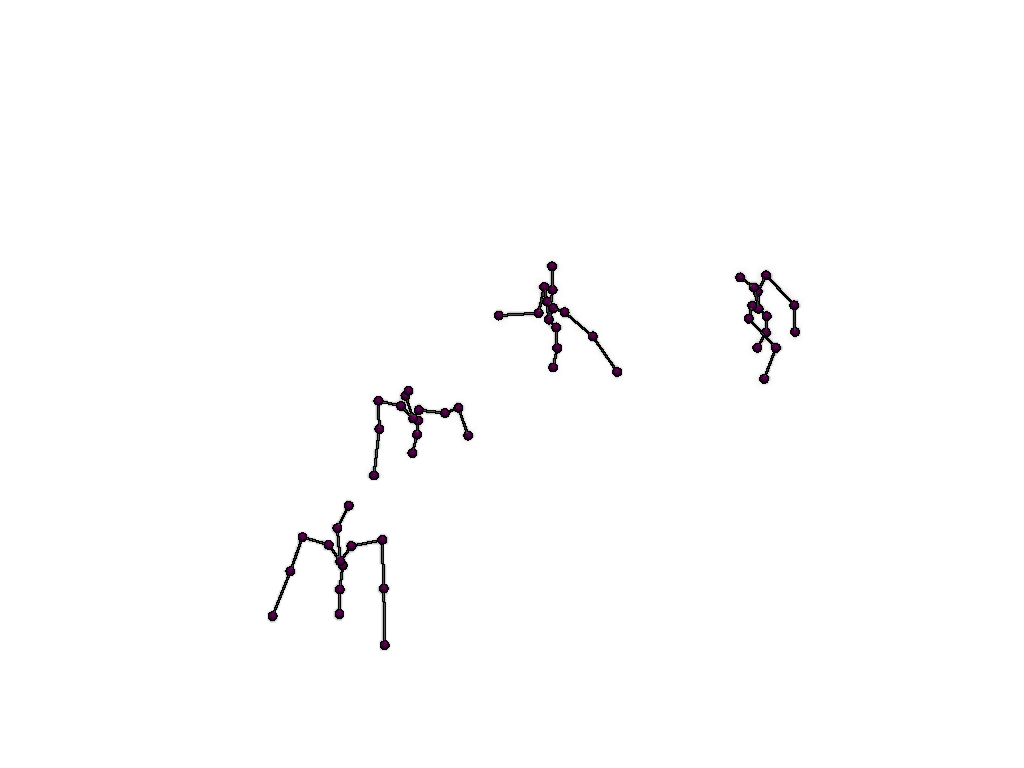} &
        \includegraphics[width=0.3\columnwidth]{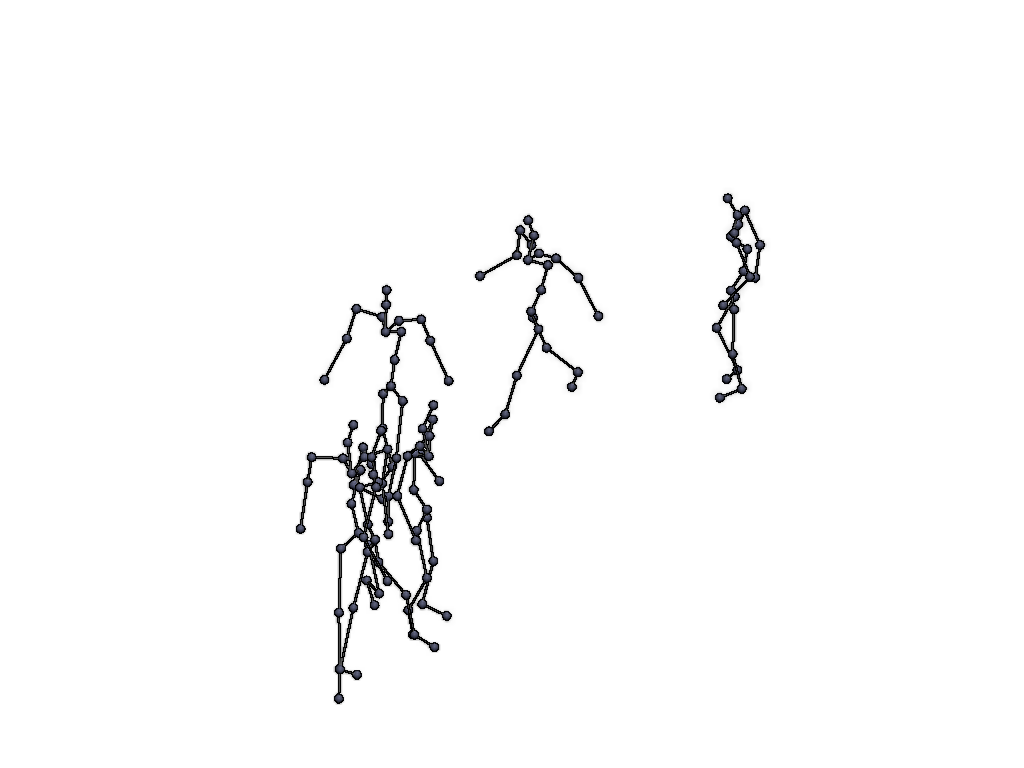} &
        \includegraphics[width=0.3\columnwidth]{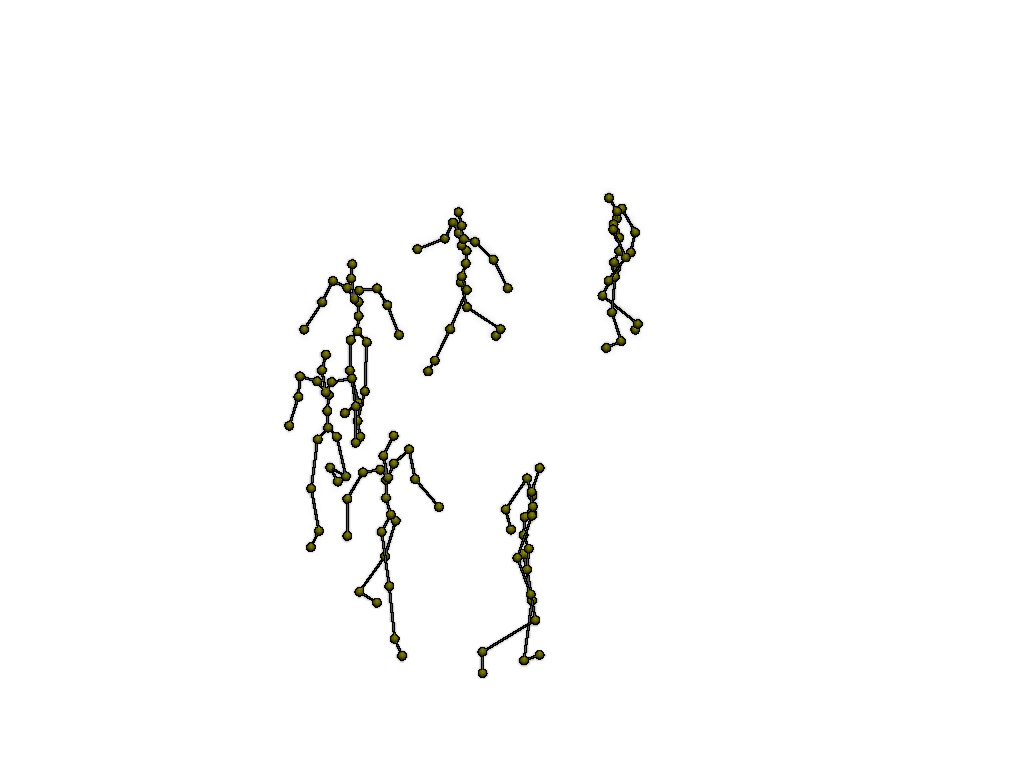} \\
        
        \includegraphics[width=0.3\columnwidth]{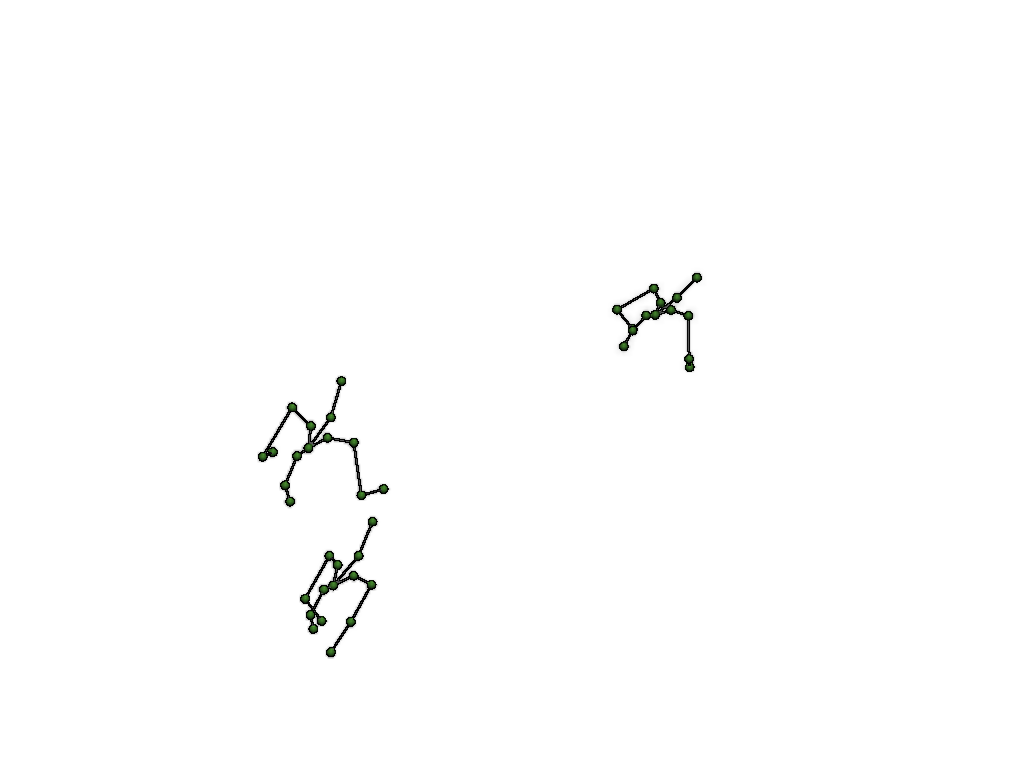} &
        \includegraphics[width=0.3\columnwidth]{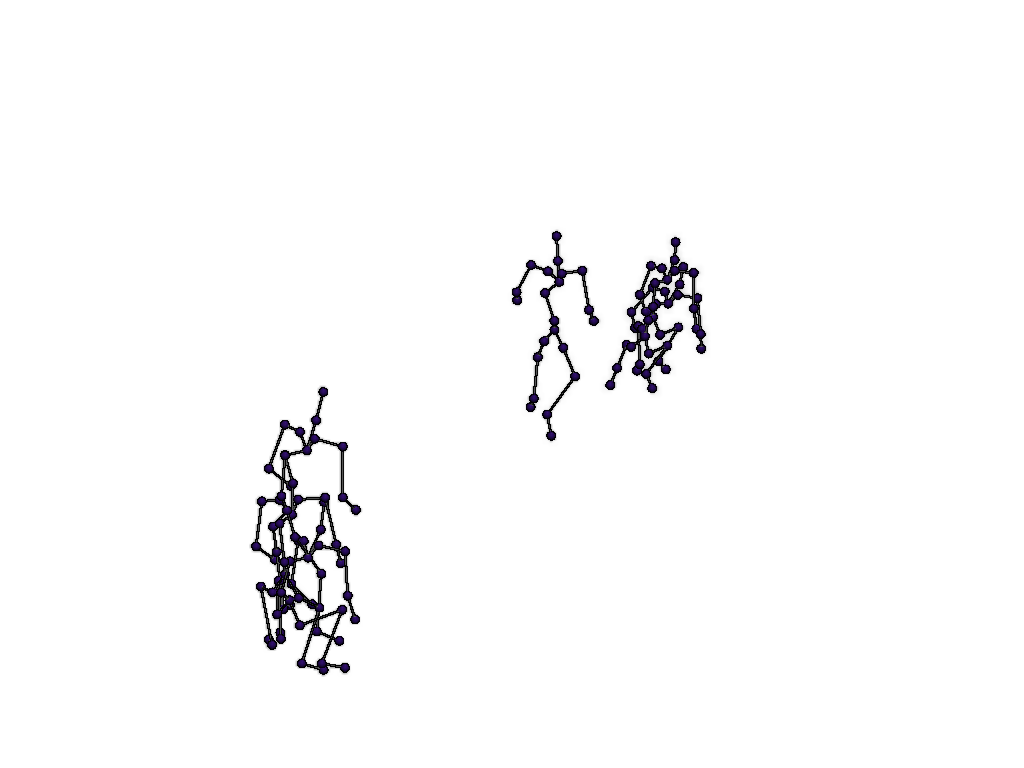} &
        \includegraphics[width=0.3\columnwidth]{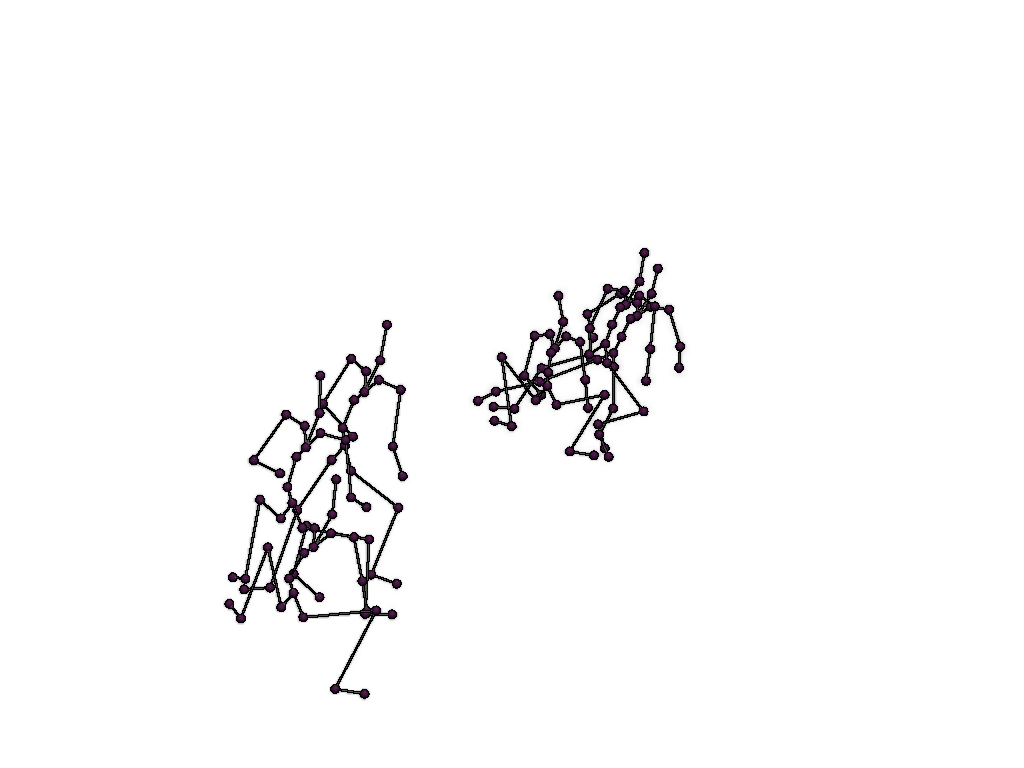} \\

        \includegraphics[width=0.3\columnwidth]{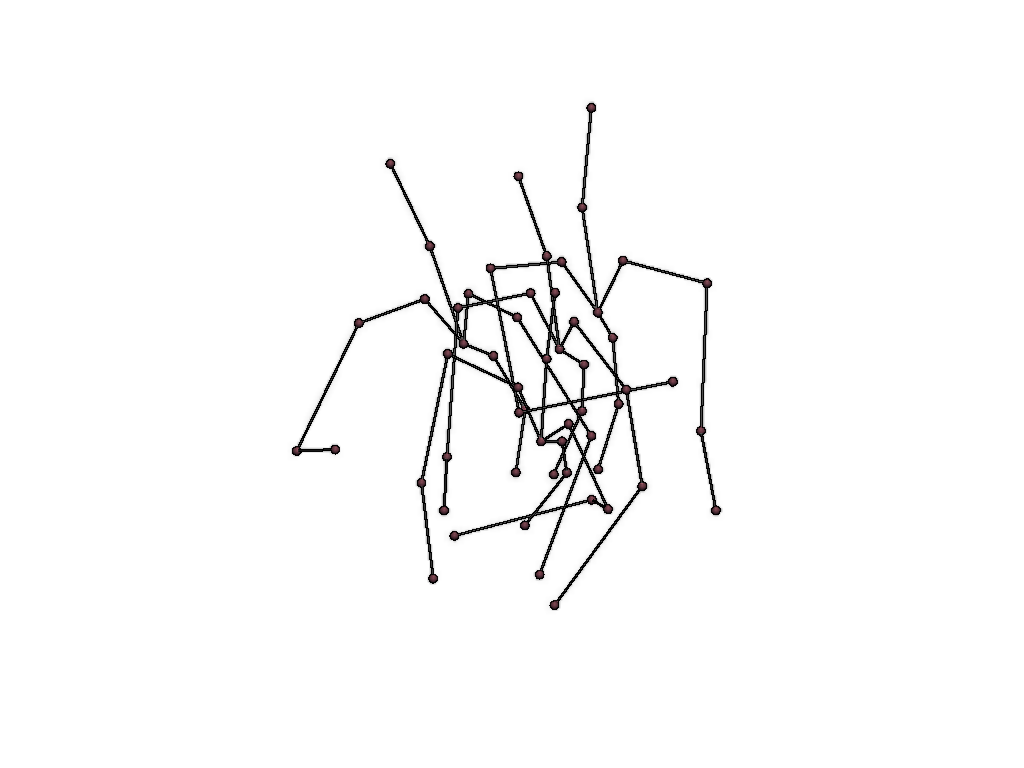} &
        \includegraphics[width=0.3\columnwidth]{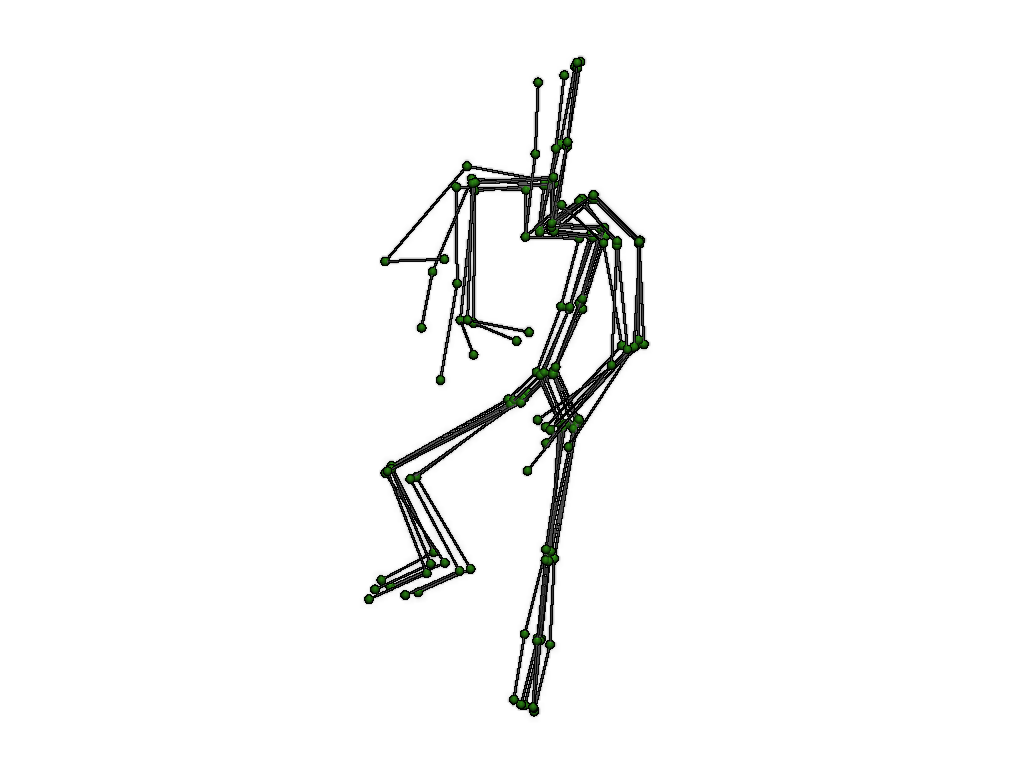} &
        \includegraphics[width=0.3\columnwidth]{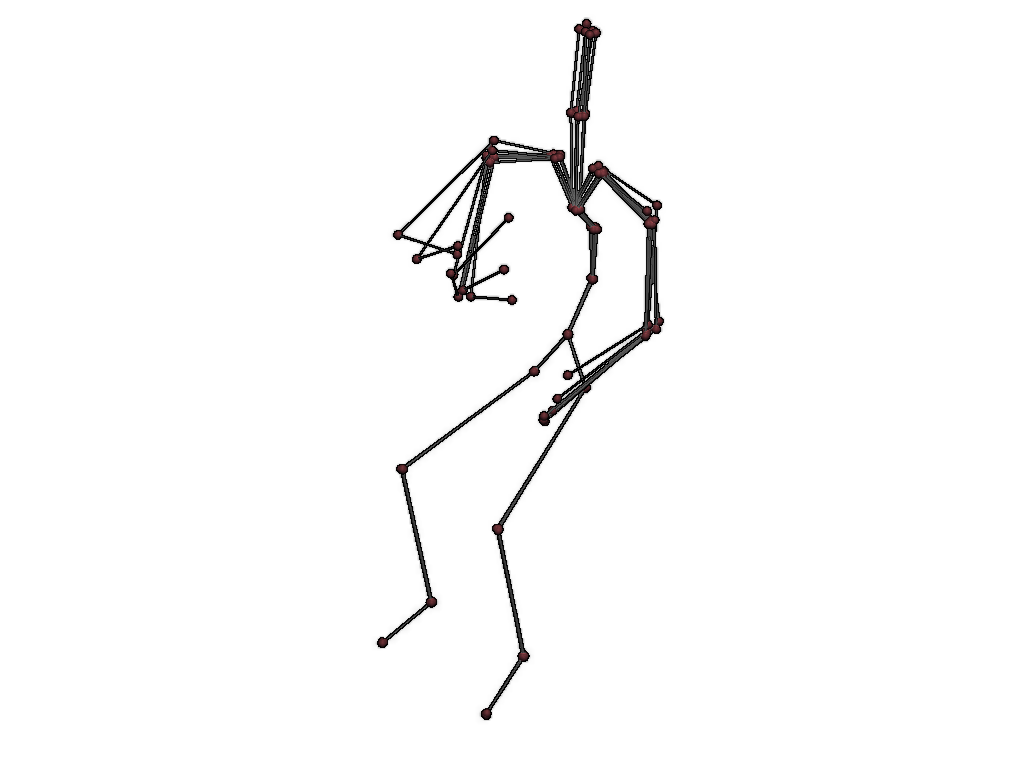} \\
        
        \includegraphics[width=0.3\columnwidth]{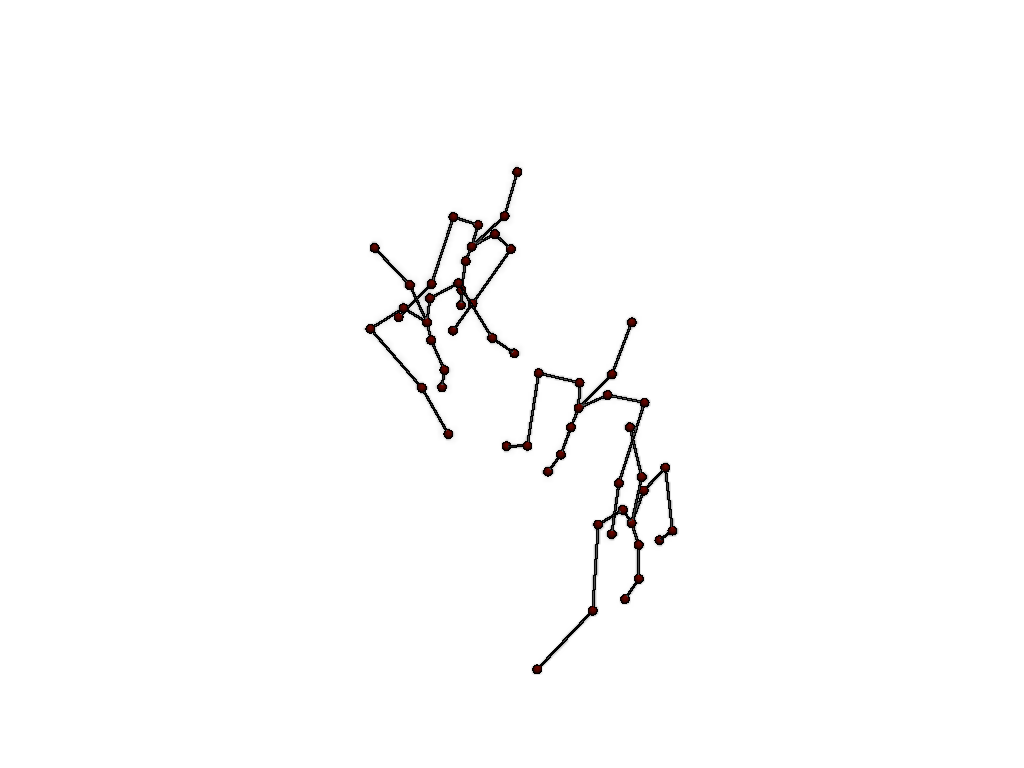} &
        \includegraphics[width=0.3\columnwidth]{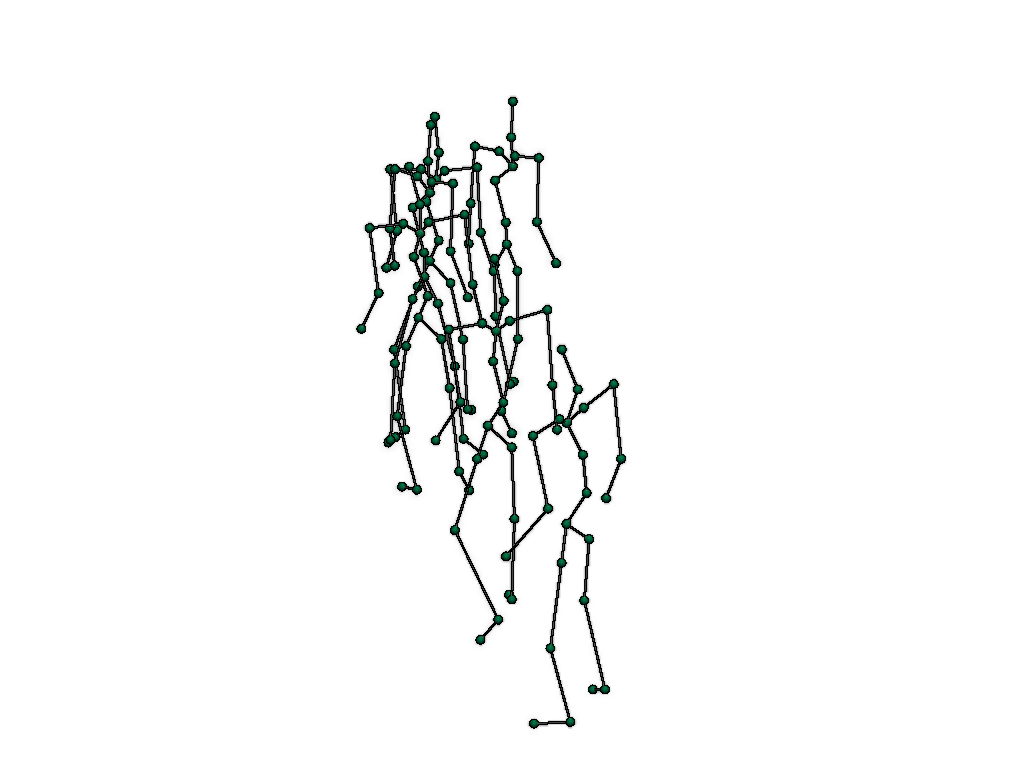} &
        \includegraphics[width=0.3\columnwidth]{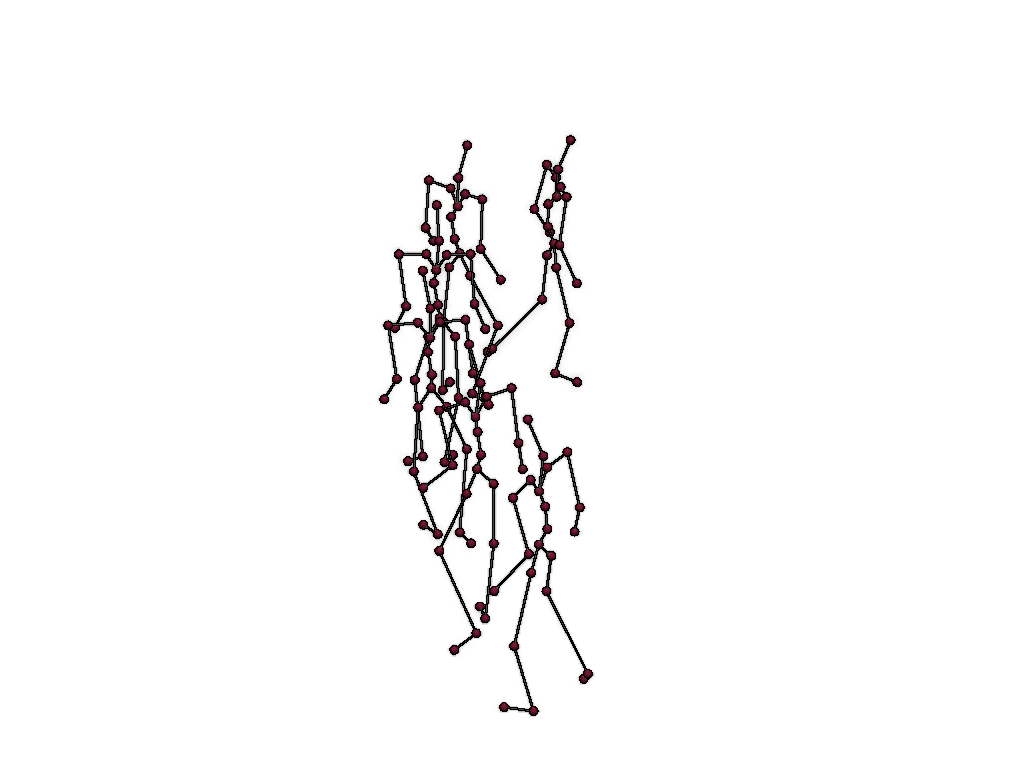} \\

        \includegraphics[width=0.3\columnwidth]{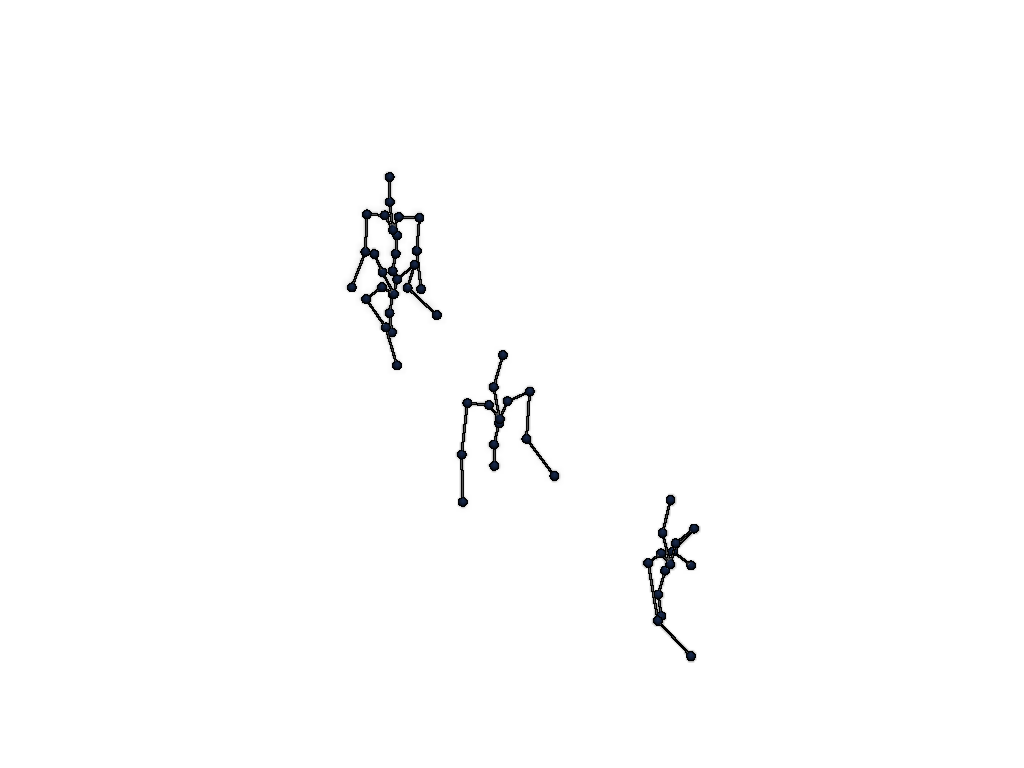} &
        \includegraphics[width=0.3\columnwidth]{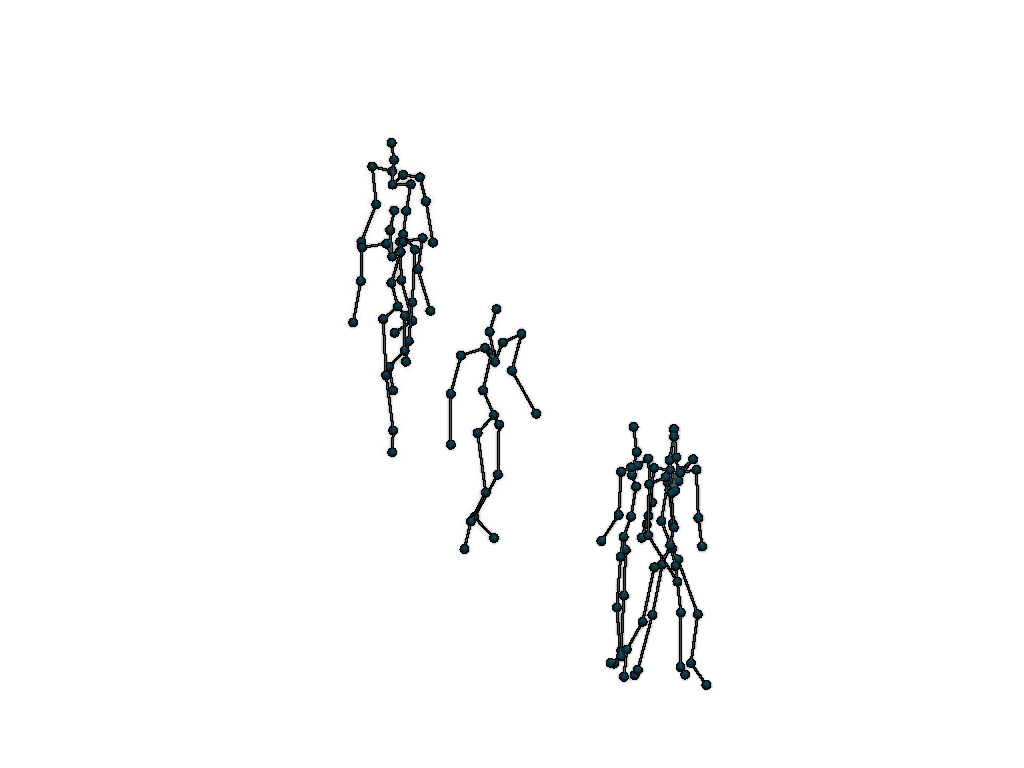} &
        \includegraphics[width=0.3\columnwidth]{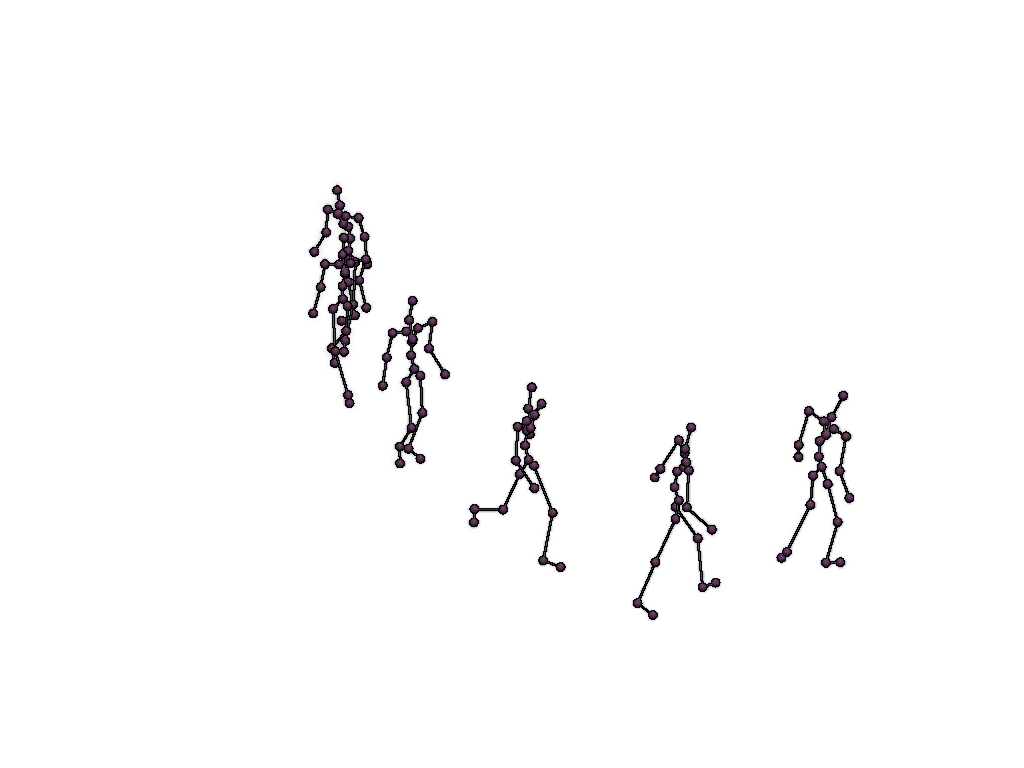} \\
        
        \includegraphics[width=0.3\columnwidth]{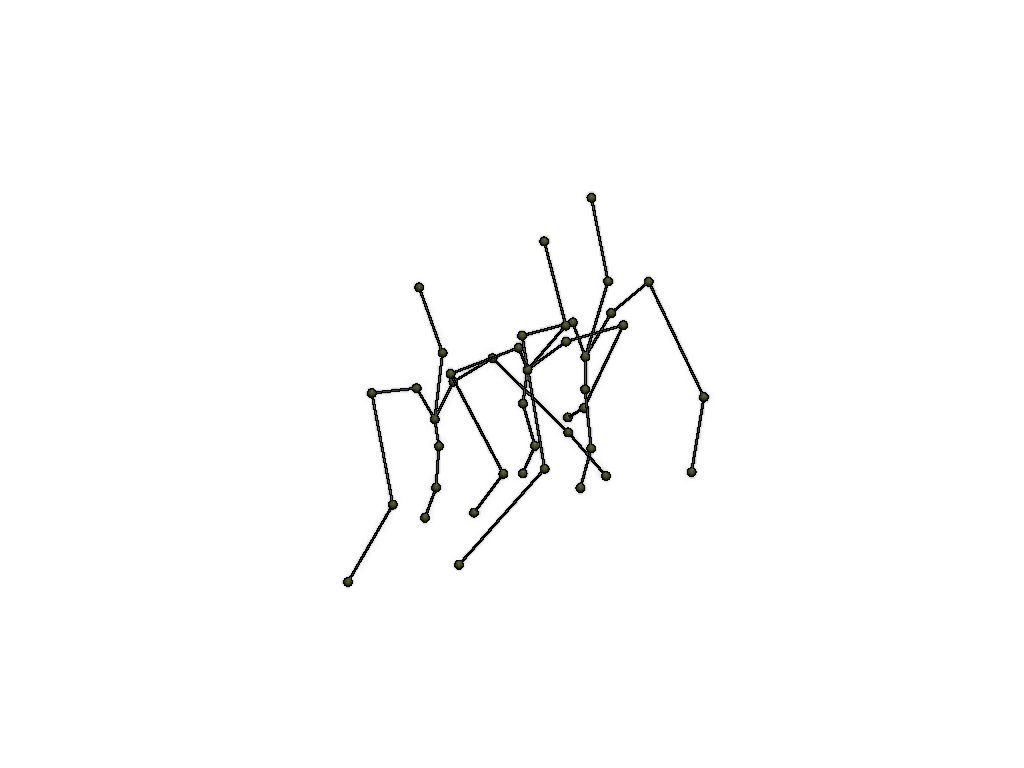} &
        \includegraphics[width=0.3\columnwidth]{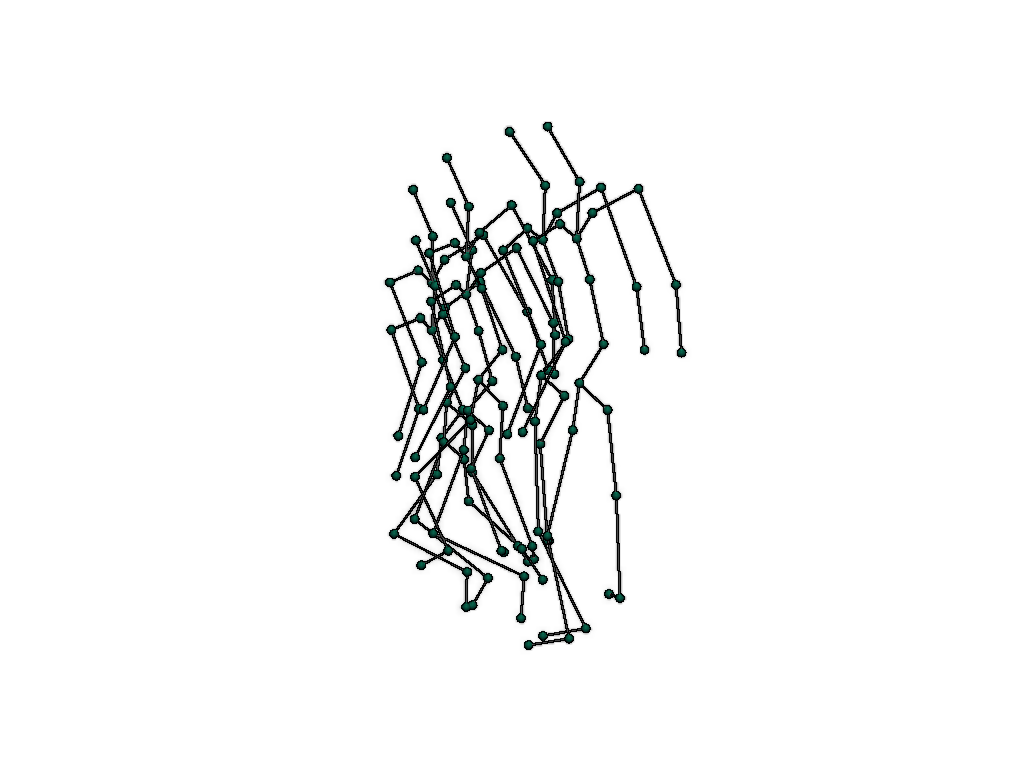} &
        \includegraphics[width=0.3\columnwidth]{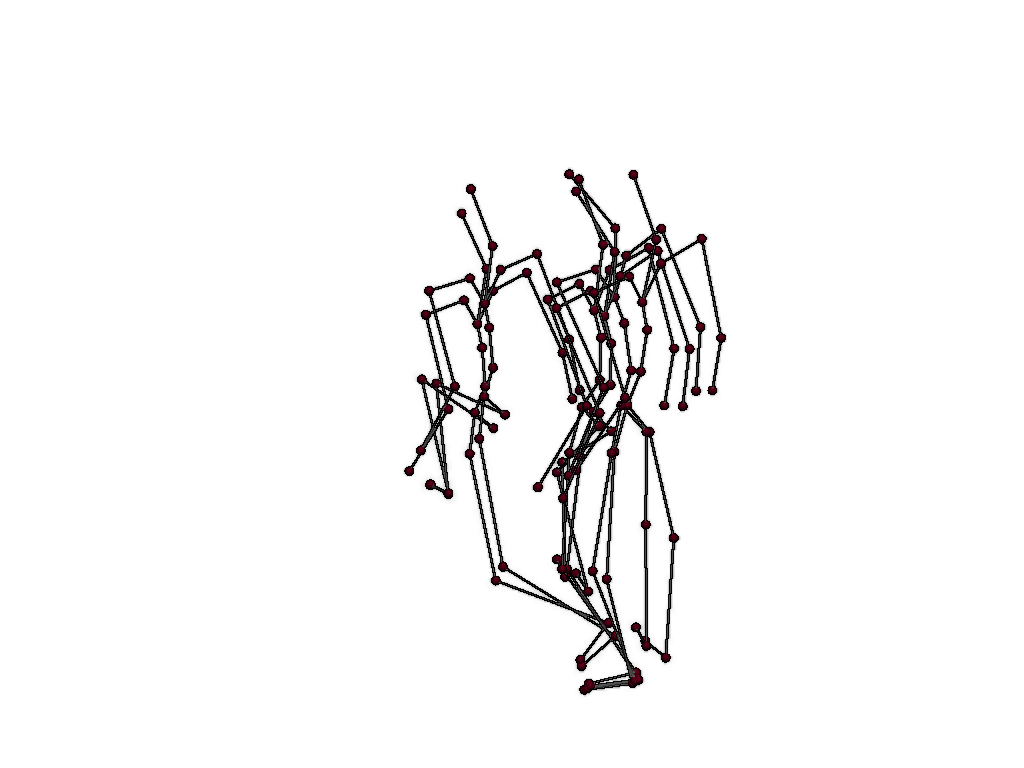} 

    \end{tabular}
    \caption{Qualitative evaluation of the motion tracking model across diverse motion types. The left column shows incomplete and noisy input observations, while the middle and right columns show samples of the reconstructed motion sequences predicted by our model.}
    \label{fig:tracking-examples-supplementary}
    \vspace{-0.8cm}
\end{figure}

\subsection{MAV Policy Evaluation}
As commonly done in social navigation, we evaluate our method on goal-directed \ac{MAV} navigation, where it must plan a safe trajectory from an initial state to a goal while avoiding human bodies as dynamic obstacles. We replay 50 AMASS~\cite{naureen2019amass} sequences in simulation as ground-truth human motion. Initial states and goals are sampled uniformly from feasible regions, and a planning horizon of 20 steps with a time delta of $0.1s$ is used. For training the policy, we use 10,000 ground-truth trajectories, that have been computed as described in \cref{seq:policy}.

\textbf{Metrics.} 
We assess performance in terms of safety (Collision Avoid.) and efficiency (Time-To-Goal). Safety is measured as the percentage of trajectories maintaining a minimum 0.5 m distance from any human joint. Efficiency captures the elapsed time to reach the goal, penalizing overly conservative behavior.

\textbf{Baselines.} 
We compare against two classes of baselines.
Distance-constraint (DC) methods formulate an MPC-like optimization problem that minimizes a goal-reaching objective while enforcing minimum-distance constraints between the \ac{MAV} and human joints along a rolled-out trajectory. These methods differ in their assumptions about human motion: (i) \textit{Static}, which fixes human joints to their last observed positions; (ii) \textit{Constant-velocity}, which extrapolates joint motion using an ORCA-inspired instantaneous velocity model~\cite{vandenberg2011orca}; and (iii) \textit{Forecast-based}, which uses predicted future human poses from our motion forecasting model.

Set-based approaches first estimate reachable sets for both the human and the \ac{MAV}. The \textit{Forward Reachable Set} baseline enforces non-overlap between the two reachable sets over the planning horizon, resulting in conservative behavior. 

Finally, we evaluate ablations of our approach. \textit{Ours + Past Joints} replaces the pretrained encoder with past human joint positions expressed in the robot frame while keeping the same flow-matching policy capacity. \textit{Ours + Past \& Future Joints} first forecasts a single future human motion rollout and provides it as additional input to the policy network. \textit{Ours + Full Avoidance} follows the same encoder-based conditioning as our method, but instead of constraining a collision-free trajectory per sample during ground-truth generation, only a single $\mathbf{u}_{opt}$ is optimized to avoid all body motion forecasts. 

\textbf{Results.}
Table~\ref{tab:policy-evaluation} summarizes navigation performance under both noiseless and noisy human motion inputs (noise level 5, as defined in \cref{seq:human-motion-eval}, e.g., regressed by a lightweight human pose estimation model).

Distance-constraint (DC) methods often produce time-efficient trajectories but do not guarantee collision-free navigation. Set-based approaches improve safety by accounting for uncertainty in future human motion, but typically rely on conservative over-approximations that lead to less time-efficient trajectories. Both distance- and set-based methods are particularly sensitive to noisy human motion estimates and may not converge in time, resulting in increased time-to-goal and higher collision rates.

Our ablation studies show that conditioning the policy directly on past joint observations or on a single forecasted trajectory improves navigation efficiency but remains insufficient to ensure safety. In contrast, conditioning the policy on our human motion model's latent representation consistently achieves 100\% collision avoidance under both clean and noisy conditions, while maintaining the lowest time-to-goal among all safe methods. Additionally, learning a policy that constrains only the initial control input for safety further improves time efficiency without compromising collision avoidance.

\textbf{Inference Details.} 
We follow standard practice in diffusion-based motion (e.g., \cite{tevet2023mdm}, [59]) and policy models (e.g., \cite{chen2025vidbot, chi2023diffusionpolicy, julbe2025diffusionmpc}). Reducing denoising steps at inference significantly lowers runtime while maintaining quality, which is well-suited to real-time settings where runtime matters more than generation diversity. In the noiseless setting, 10 steps achieve nearly identical performance to 20 steps (100\% collision avoidance, 3.80s vs.\ 3.79s time-to-goal), while fewer steps degrade safety (98\% collision avoidance at 7 steps). At 3 steps, we get 100\% collision avoidance but with a longer 4.45s time-to-goal. We observe similar trends under noisy conditions (noise level = 5), with 10 steps maintaining 100\% collision avoidance. Based on this trade-off between efficiency and consistently safe behavior, we use 10 steps across all experiments in the final setting.

\textbf{Distribution Shift.}
We evaluate under complete distribution shift using dancing motion from the AIST++ dataset~\cite{li2021aist}, which neither the human motion model nor the control policy has seen during training. As shown in \cref{tab:policy-evaluation-w-distribution}, the system achieves 100\% collision avoidance with a median time-to-goal of 3.15s on AIST++ dance sequences, demonstrating effective generalization to unseen scenarios without online adaptation. The compressed latent representation captures motion primitives that transfer across motion types.

\begin{table}[ht!]
\centering
\begin{tabular}{l | c | ccc}
\toprule
\multirow{2}{*}[-0.3em]{\textbf{Method}}
& \multirow{2}{*}[-0.3em]{\makecell{\textbf{Collision}\\\textbf{Avoid. $\uparrow$ [\%]}}}
& \multicolumn{3}{c}{\textbf{Time-to-Goal $\downarrow$ [s]}} \\
\cmidrule(l){3-5}
& & \textbf{25th} & \textbf{50th} & \textbf{75th} \\
\midrule
& \multicolumn{4}{c}{\textbf{No Noise}} \\
\midrule
HumanHalo~[45] & \textbf{100} & 3.53 & 4.46 & 5.33 \\
Ours & \textbf{100} & \textbf{3.10} & \textbf{3.80} & \textbf{5.10} \\
\midrule
& \multicolumn{4}{c}{\textbf{Noise Level = 5}} \\
\midrule
HumanHalo~[45] & 92 & 3.46 & 5.47 & 6.41 \\
Ours & \textbf{100} & \textbf{3.05} & \textbf{3.91} & \textbf{4.80} \\
\midrule
& \multicolumn{4}{c}{\textbf{AIST++ Dataset}} \\
\midrule
Ours & 100 & 1.13 & 3.15 & 6.08 \\
\bottomrule
\end{tabular}
\caption{Quantitative simulation results of the MAV policy including time-to-goal distribution (25th, 50th, and 75th percentiles) and out-of-distribution dataset AIST++~\cite{li2021aist}.}
\label{tab:policy-evaluation-w-distribution}
\vspace{-0.7cm}
\end{table}

\section{Limitations}
\label{sec:limitations}

\textbf{Human Motion Modeling.}
While our model captures full-body motion and human–scene interactions, it does not explicitly account for social interactions between multiple humans, where individuals influence each other’s motion. Additionally, hand motion is not modeled; incorporating hand dynamics could further improve reasoning about fine-grained scene interactions such as object manipulation or support contacts.

\textbf{Modeling Human--Robot Interactions.}
The \ac{MAV} policy is trained primarily for collision avoidance rather than explicit human–robot interaction, due to the limited availability of data involving close human–\ac{MAV} encounters. Also, our framework assumes that human motion is unaffected by the \ac{MAV}’s behavior. While this assumption simplifies planning, humans may react to nearby robots and adapt their motion. Modeling such bidirectional human–robot interactions has only been studied for 2D root trajectory generation \cite{salzmann2020trajectron++} and remains an open challenge for full 3D body motion. Lastly, we use a simplified linear drone dynamics model for control. Although this enables efficient optimization for generating ground-truth \ac{MAV} trajectories, our framework allows for efficient generation of controls independent of the underlying \ac{MAV} dynamics. Thus, extending the policy to more realistic or higher-order dynamics could improve flight agility.

\textbf{Data Scale and Diversity.}
Although our method generalizes well across the evaluated benchmarks, datasets capturing human–scene interactions remain limited in scale and diversity. In particular, GIMO~\cite{zheng2022gimo} includes only a small number of indoor scenes. Expanding training data to more varied environments, including outdoor settings, would likely further improve robustness.

\section{Conclusion} 
\label{sec:conclusion}
We introduced \algname, a latent diffusion model for 3D scene-conditioned joint human motion tracking and forecasting. We further demonstrated its effectiveness in robotic applications by integrating it into a 3D \ac{MAV} social navigation policy. In particular, we showed how the latent representations learned by \algname{} can be leveraged within a flow-matching-based control policy, enabling collision-free and computationally efficient control, even under noisy and occluded observations.

Extensive experiments demonstrate that our human motion model produces smooth and accurate predictions, outperforming state-of-the-art tracking methods at a fraction of their computational cost. When integrated into the navigation pipeline, our approach enables consistently safe and efficient \ac{MAV} control. The system remains collision-free even under severe noise and partial observability, outperforming strong optimization- and reachability-based baselines.

Our results highlight the importance of jointly reasoning about human motion uncertainty and robot control within a shared latent representation. Future work will focus on extending the framework to richer human–robot interaction scenarios and scaling to more diverse and unstructured environments. While we demonstrate \algname{} in the context of \ac{MAV} navigation, the underlying formulation is general and applicable to a broader range of robotic systems, representing a step toward robust, human-aware robotic operation in real-world environments.

\section*{Acknowledgments}
This work was supported by TUM Georg Nemetschek Institute under the VAULT-AI project, by TUM AGENDA 2030, funded by the Federal Ministry of Education and Research and the Free State of Bavaria, and by the ETH RobotX research grant funded through the ETH Zurich Foundation. We thank anonymous reviewers for their constructive comments.

\clearpage
\bibliographystyle{plainnat}
\bibliography{references}

\clearpage
\setcounter{page}{1}
\setcounter{section}{0}
\renewcommand\thesection{\Roman{section}}

\twocolumn[{%
\begin{center}
    {\Huge \algname{} -- Supplementary Material}
\end{center}
\vspace{1cm}
}]

\section{Model Architectures}
\subsection{Human Motion Model}
\textbf{Autoencoder.}
Before encoding, the input is flattened along the joint dimension and processed by two temporal downsampling convolutional blocks. Each block consists of a 1D convolution with kernel size 3 and stride 2, followed by ReLU~\cite{agarap2019relu}, and a 1D convolution with kernel size 1 followed by ReLU~\cite{agarap2019relu}. This reduces the temporal resolution by a factor of four and produces latent features with dimensionality 128. Sinusoidal positional encodings are added and the sequence is processed by a 6-layer transformer encoder with 8 attention heads, feed-forward dimension four times the model dimension, and dropout 0.1. The noisy encoder $E_{B}(\cdot)$ follows a similar architecture.
The latent representation is processed by the decoder $D$, a transformer encoder with the same configuration as in $E$. Temporal resolution is restored using two upsampling convolutional blocks. Each block performs nearest-neighbor upsampling by a factor of two, followed by a 1D convolution with kernel size 3 and ReLU~\cite{agarap2019relu}, and a 1D convolution with kernel size 1 and ReLU~\cite{agarap2019relu}. A final upsampling layer followed by a 1D convolution projects the features to the output dimension, producing reconstructed motion with full temporal length.

\textbf{Scene Encoder.} 
We encode the scene occupancy grid using a 3D convolutional encoder. The encoder consists of three 3D convolutional layers with channel dimensions $1 \rightarrow 8 \rightarrow 16 \rightarrow 32$, each using kernel size 3, padding 1, and ReLU activation. The resulting features are aggregated using global average pooling to obtain a single feature vector per grid. A linear layer then projects the features to a latent embedding of dimension 128.

\textbf{Diffusion Model.}
We use a Transformer-based denoising network operating on latent motion representations. The model conditions on diffusion timesteps and encoded context features. Diffusion timesteps are embedded using a learned timestep embedding and added together with sinusoidal positional encodings.
The input latent representation is first projected to a hidden dimension of 256 using a linear layer. The denoising network consists of a 10-layer Transformer encoder with FiLM~\cite{perez2018film} conditioning for the scene, where noisy body latents are concatenated to the input noise. Each layer uses 8 attention heads, feed-forward dimension four times the hidden dimension, GELU~\cite{hendrycks2016gelu} activation, and dropout 0.1. The output is projected back to the latent dimensionality using a linear layer.

\subsection{MAV Policy Model}
The flow-matching-based \ac{MAV} policy model is implemented as a \ac{MLP} conditioned on encoded context features. Conditioning sequences are processed by a three-layer MLP with dimensions $\text{input\_dim} \rightarrow 64 \rightarrow 32 \rightarrow 8$ with ReLU~\cite{agarap2019relu} activations, and flattened across time.
The processed latents are concatenated with the input noise, the encoded features for the goal position and initial state as well as the denoising timestep. The final prediction network consists of an \ac{MLP} with layer dimensions inputs $\rightarrow 256 \rightarrow 128 \rightarrow 64 \rightarrow 3T$, where $T$ is the prediction horizon. Each hidden layer is followed by a ReLU~\cite{agarap2019relu} activation. The network outputs control sequences matching the prediction horizon.

\section{Uncertainty Estimates of Reported Metrics}
We report mean $\pm$ standard deviation for the AMASS~\cite{naureen2019amass} Occ-10\% tracking benchmark in \cref{tab:tracking-amass-pm}, computed over the AMASS~\cite{naureen2019amass} evaluation sequences. Our method achieves both the lowest error and tightest bounds across all metrics, confirming robustness.

\begin{table}[!ht]
\centering
\scriptsize
\begin{tabular}{c | c | c c c c}
\toprule
Input & Noise & GMPJPE-all $\downarrow$ & Accel $\downarrow$ & Skating $\downarrow$ & Dist $\downarrow$ \\
\midrule
\multirow{3}{*}{Occ-L} & 3 & 43.78 {$\pm$ 4.4} & 0.58 {$\pm$ 0.01} & 0.21 {$\pm$ 0.09} & 1.55 {$\pm$ 0.5} \\
& 5 & 50.99 {$\pm$ 6.1} & 0.60 {$\pm$ 0.02} & 0.17 {$\pm$ 0.09} & 1.42 {$\pm$ 0.6} \\
& 7 & 65.0 {$\pm$ 7.8} & 0.64 {$\pm$ 0.04} & 0.14 {$\pm$ 0.07} & 1.00 {$\pm$ 0.5} \\
\midrule
Occ-10\% & 3 & 19.8 {$\pm$ 5.7} & 0.48 {$\pm$ 0.01} & 0.10 {$\pm$ 0.04} & 0.52 {$\pm$ 0.2} \\
\bottomrule
\end{tabular}
\caption{Scene-Free human motion tracking evaluation on AMASS~\cite{naureen2019amass} with uncertainty estimates.}
\label{tab:tracking-amass-pm}
\vspace{-0.5cm}
\end{table}

\section{Model and Loss Function Hyperparameters.}
The hyperparameters for the human motion model architecture and the associated loss weights were selected via grid search over a held-out validation split. As an example, \cref{tab:tracking-amass-ours-occ10} reports an ablation study quantifying the effect of different loss weight settings on model performance, demonstrating the robustness of our chosen configuration.

\begin{table}[ht!]
\centering
\scriptsize
\begin{tabular}{c c | c c c c}
\toprule
$\lambda_{\text{vel}}$ & $\lambda_{\text{bone}}$ & GMPJPE-all $\downarrow$ & Accel $\downarrow$ & Skating $\downarrow$ & Dist $\downarrow$ \\
\midrule
0.1 & 0.01 & 19.8 {$\pm$ 5.7} & 0.48 {$\pm$ 0.01} & 0.10 {$\pm$ 0.04} & 0.52 {$\pm$ 0.20} \\
0.1 & 0.1 & 22.1 {$\pm$ 8.8} & 0.42 {$\pm$ 0.02} & 0.11 {$\pm$ 0.05} & 0.69 {$\pm$ 0.29} \\
0.01 & 0.001 & 23.2 {$\pm$ 9.2} & 0.54 {$\pm$ 0.01} & 0.15 {$\pm$ 0.04} & 1.40 {$\pm$ 0.40} \\
\bottomrule
\end{tabular}
\caption{Ablation of human motion tracking on AMASS~\cite{naureen2019amass} for the Occ-10\% setting across loss weight parameters.}
\label{tab:tracking-amass-ours-occ10}
\vspace{-0.5cm}
\end{table}

\end{document}